\documentclass[10pt,twocolumn,letterpaper]{article}

\usepackage[pagenumbers]{iccv} 
\usepackage{multirow}
\usepackage{tabularx}
\usepackage{graphicx}
\usepackage{textcomp}
\usepackage{marvosym}

\definecolor{iccvblue}{rgb}{0.21,0.49,0.74}
\usepackage[pagebackref,breaklinks,colorlinks,allcolors=iccvblue]{hyperref}

\def\paperID{7562} 
\def\confName{ICCV}
\def\confYear{2025}

\title{IDMR: Towards Instance-Driven Precise Visual Correspondence in Multimodal Retrieval}

\author{Bangwei Liu$^{1*}$
\qquad
    Yicheng Bao$^1$  \qquad
    Shaohui Lin$^1$ \\
    Xuhong Wang$^2$  \textsuperscript{\Letter} \qquad
    Xin Tan$^{1}$ \textsuperscript{\Letter} \qquad
    Yingchun Wang$^2$ \qquad
    Yuan Xie$^1$ \qquad
    Chaochao Lu$^{2}$ \\
    {\normalsize$^1$East China Normal University} \qquad
    {\normalsize$^2$Shanghai AI Laboratory} \\
    {\tt\small 51265901118@stu.ecnu.edu.cn, wangxuhong@pjlab.org.cn, xtan@cs.ecnu.edu.cn}
}

\begin{document}
\maketitle
\begin{abstract}

Multimodal retrieval systems are becoming increasingly vital for cutting-edge AI technologies, such as embodied AI and AI-driven digital content industries. However, current multimodal retrieval tasks lack sufficient complexity and demonstrate limited practical application value. 
It spires us to design Instance-Driven Multimodal Image Retrieval (IDMR), a novel task that requires models to retrieve images containing the same instance as a query image while matching a text-described scenario. Unlike existing retrieval tasks focused on global image similarity or category-level matching, IDMR demands fine-grained instance-level consistency across diverse contexts. To benchmark this capability, we develop IDMR-bench using real-world object tracking and first-person video data. Addressing the scarcity of training data, we propose a cross-domain synthesis method that creates 557K training samples by cropping objects from standard detection datasets. Our Multimodal Large Language Model (MLLM) based retrieval model, trained on 1.2M samples, outperforms state-of-the-art approaches on both traditional benchmarks and our zero-shot IDMR-bench. Experimental results demonstrate previous models' limitations in instance-aware retrieval and highlight the potential of MLLM for advanced retrieval applications. The whole training dataset, codes and models, with wide ranges of sizes, are available at \href{https://github.com/BwLiu01/IDMR}{https://github.com/BwLiu01/IDMR}.
\end{abstract}


\section{Introduction}
\label{sec:intro}

\begin{figure}[t]
  \centering
   \resizebox{0.75\linewidth}{!}{
       \includegraphics[width=0.95\linewidth]{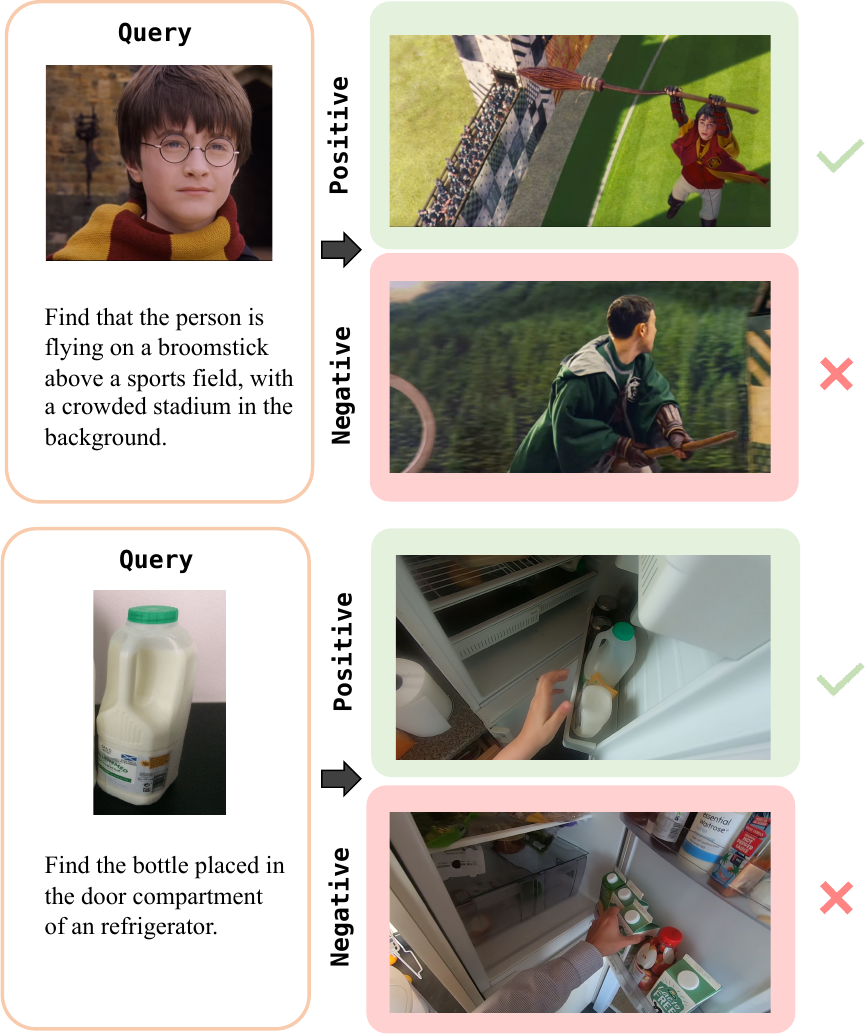}
   }
   \caption{Examples of Instance-driven Multimodal Image Retrieval task. The positive image should contain the same instance as in the query and also comply with the query text. The negative image only adheres to the query text but does not contain the same instance present in the query.  }
   \vspace{-5mm}
   \label{fig:retrieve example in intro}
\end{figure}


Multimodal retrieval systems play a pivotal role in boosting tasks such as information retrieval \cite{Faysse2024ColPaliED, li2024matching_information_retrieval} and Retrieval Augmented Generation (RAG) \cite{qian2024memorag, packer2023memgpt_rag, salemi2024evaluating_rag}.  
The emergence of frontier AI technologies, exemplified by multimodal large language models, has heightened the demand for sophisticated multimodal data retrieval techniques. As illustrated in~\cref{fig:retrieve example in intro}, embodied robots powered by multimodal models should precisely access their image memory repositories to recreate similar scenarios, thereby enhancing their execution of user directives. The exponential growth of AI-generated and human-recorded video content demands sophisticated indexing technologies capable of mapping specific individuals to relevant scenes—a critical foundation for enabling MLLM-powered automated video production pipelines in the future.

\begin{figure}[t]
  \centering
       \includegraphics[width=1\linewidth]{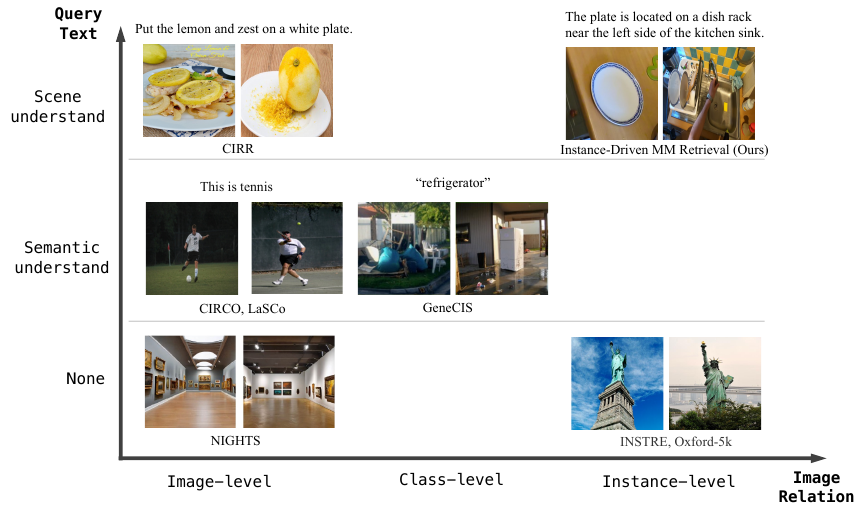}
   \caption{Comparison of IDMR-bench with other benchmarks. We focus on Instance-level retrieval and query describing complex scenes. Zoom in best.}
   \label{fig:retrieve example compare}
\end{figure}

This insight has inspired our proposal of a novel {\bf I}nstance-{\bf D}riven {\bf M}ultimodal {\bf R}etrieval ({\bf IDMR}) task, with heightened focus on a critical scenario: given a query image of an instance and a query text describing the scenario in which this instance appears, the task is to find an image within a large image database that contains the same instance and matches the scenario described by the query text. The key improvement between the IDMR task and other previous tasks is shown in \cref{fig:retrieve example compare}. 

Single-modal retrieval benchmarks like NIGHTS~\cite{fu2023dreamsim_nights} emphasize high similarity across the entire image, while INSTRE~\cite{Wang2015INSTRE} focuses on instance-level consistency. However, INSTRE does not incorporate a text modality, which limits its ability to execute richer query intents.
Composed Image Retrieval (CIR) benchmarks like CIRR~\cite{cirr} assume that the relationship between the query and the positive sample is based on global image similarity, neglecting fine-grained instance-level correspondence. GeneCIS \cite{genecis} takes a step further by considering image similarity based on consistent object categories and attributes. It defines images with matching object classes and attributes as similar. However, objects with the same semantic category and attributes do not necessarily belong to the same instance. 

Although our proposed new task holds promise to advance the application of multimodal retrieval, neither traditional computer vision techniques such as object detection~\cite{objects365} nor visual-language alignment methods like CLIP~\cite{clip} can technically achieve this fine-grained visual-language alignment retrieval functionality. 
Recent advances in multimodal embedding models \cite{magiclens, e5v, vlm2vec, gme} have shown promising results on benchmarks like cross-modal retrieval~\cite{coco, flickr30k, fashion200k} and fused-modal retrieval \cite{cirr, circo, lasco}. However, these benchmarks focus primarily on global image alignment, emphasizing holistic semantic similarity rather than fine-grained instance-level details. 

In this work, we introduce a novel Instance-Driven Multimodal Image Retrieval (IDMR) task and develop a benchmark named IDMR-bench to assess whether models can retrieve images of the same instance across diverse contexts. Specifically, we compile real-world object tracking and first-person view \textbf{video} as benchmarking data, which provide continuous visual recordings of a particular instance over varying frames and scenes. However, the scarcity of such benchmarking data makes scaling up the training dataset challenging. To address this, we propose a cross-domain training data synthesizing method that simply cropping an object from an image and using it as the image query. This approach allows us to synthesize 557K large-scale training samples using three commonly used object detection \textbf{image} as training datasets, while while maintaining the potential for continuous scaling.
Together with existing multimodal retrieval datasets, we totally used 1.2M data to train an MLLM-based multimodal retrieval model.

Experiments among three kinds of benchmarks, including a widely used MMEB~\cite{vlm2vec} and our zero-shot IDMR-bench, demonstrate that current state-of-the-art models~\cite{vlm2vec,lin2024mm-embed} in multimodal retrieval exhibit limitations, suggesting a gap in their capacity to accurately identify and retrieve identical instances. Our model outperforms the state-of-the-art (SOTA) model in almost all quantitative and qualitative studies. The performance on the MMEB benchmark shows that our model has the capability to continuously learn from traditional retrieval tasks to our more challenging IDMR task. The generalization of zero-shot IDMR-bench substantiates the formidable potential of Multimodal Large Language Models in real-world multimodal retrieval applications. 






\section{Related Work}
\subsection{Multimodal Retrieval Benchmarks}

Early multimodal retrieval benchmarks primarily focused on cross-modal retrieval \cite{coco, flickr30k, fashion200k}, where a query text is matched with candidate images based on holistic semantic similarity, or vice versa (query and candidate modalities are interchangeable). 
These retrieval tasks emphasize global image-text alignment, aiming to capture broad semantic relationships between images and texts.

With the development of Vision-Language Models (VLMs) \cite{clip,siglip, openclip,blip} and Multimodal Large Language Models (MLLMs) \cite{ liu2023llava,liu2024llavanext,qwen2-vl,abdin2024phi,internvl2.5}, the demands of fused multimodal retrieval have become achievable. 
This progress has spurred the creation of more sophisticated multimodal retrieval benchmarks, particularly for image-text to image (IT2I) retrieval.
For example, the pioneering IT2I retrieval benchmark FashionIQ\cite{fashioniq} aims to evaluate whether a model can retrieve a target clothing item given a source image of the item and a text query specifying desired attribute changes (e.g., color, style, or pattern)
However, it focuses exclusively on the fashion domain, which restricts its generalizability to broader multimodal tasks. 
Other benchmarks such as CIRR\cite{cirr}, CIRCO\cite{circo} and LaSCo\cite{lasco} focus on natural image domain and construct query data by constructing visually similar image pairs, but these pairs may lack the granularity needed to distinguish fine-grained similarities. 
Additionally, GeneCIS \cite{genecis} takes a step further by considering image similarity based on consistent object categories and attributes. It defines images with matching object classes and attributes as similar. However, this approach has a critical limitation: objects with the same semantic category and attributes do not necessarily belong to the same instance. 
To our knowledge, we are the first to propose a multimodal retrieval benchmark on instance-level consistency and complex scene understanding. 

Universal Retrieval Benchmarks like M-BEIR \cite{uniir} and UMRB \cite{gme} collect a group of evaluation datasets that cover different types of multimodal retrieval tasks including single modal retrieval, cross-modal retrieval and interleaved text-image retrieval. 
MMEB \cite{vlm2vec} includes not only retrieval tasks but also transforms image classification, Visual Question Answering (VQA), and visual grounding into retrieval tasks as part of the benchmark. To enhance the generalization of our model, we have combined the training data from MMEB and achieved performance surpassing the SOTA model on MMEB.

\subsection{Multimodal Retrieval Models}
Early Multimodal Embedding Models like CLIP \cite{clip}, BLIP \cite{blip} and CoCa \cite{coca} learn to encode images and text separately from image-text pairs. 
They achieve strong performance in aligning visual and textual representations.
Pic2Word \cite{saito2023pic2word}, SPRC \cite{sprc}, UniIR \cite{uniir} extend pretrained vision-language models to multimodal retrieval tasks by leveraging late fusion \cite{saito2023pic2word, baldrati2022effective, uniir}
and sentence-level prompting \cite{sprc} techniques. 
Recent works attempt various approaches to construct multimodal retrieval samples for training general multimodal embedding models. 
These methods range from leveraging naturally co-occurring image-text pairs to generating synthetic data or mining other image pairs with similar relationships.
For instance, MagicLens\cite{magiclens} relies on collecting naturally occurring image pairs from the same webpage to create diverse query data. Although this approach introduces variety, it often fails to ensure that the pairs exhibit meaningful visual correspondence. 
Despite these efforts, none of these methods reliably identify visually identical targets, limiting their effectiveness in real-world applications where precise visual matching is essential. 
VLM2Vec \cite{vlm2vec} and MM-Embed \cite{lin2024mm-embed} are general multimodal embedding models based on MLLMs. We adopt a similar architecture and achieve SOTA performance on the General Multimodal Embedding Benchmark.

\section{Task Definition and Benchmark}
In this section, we first provide the definition of our proposed Instance-driven Multimodal Retrieval task and introduce two distinct schemas for this novel task. 
Next, we construct a series of evaluation datasets specifically designed for zero-shot testing called IDMR-bench. 
Finally, we propose a simple yet scalable method to automatically construct instance-driven multimodal retrieval samples for training.

\begin{figure*}
  \centering
    \includegraphics[width=0.8\linewidth]{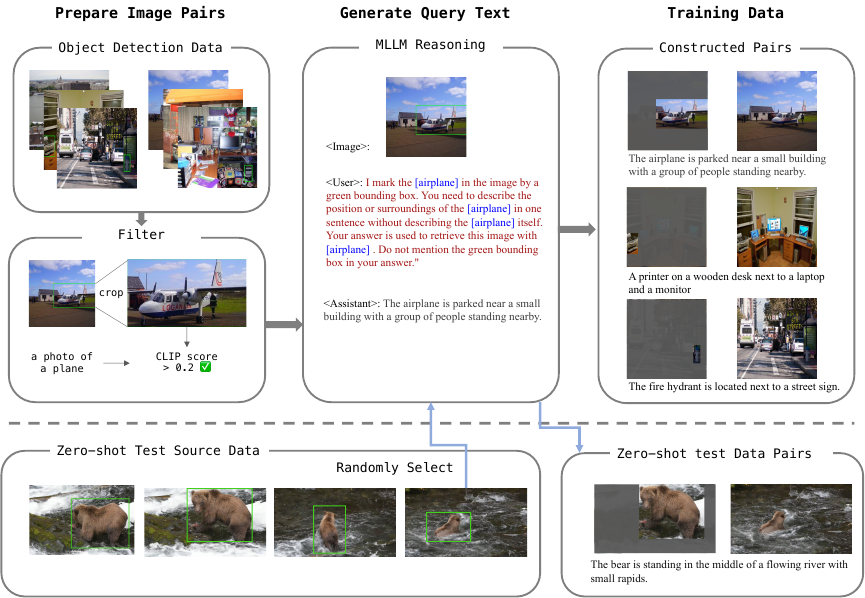}
  \caption{Data construction pipeline. The training data (top) and zero-shot benchmark data (bottom) are from different sources.}
  \vspace{-3mm}
  \label{fig:data pipeline}
\end{figure*}

\subsection{Task Definition}

We define Instance-driven Multimodal Image Retrieval (IDMR) as a task where the model is provided with a query image $I_q$ with a reference instance (e.g., an image of a bottle) and a textual description $T_q$ specifying the target context surrounding the instance (e.g., ``The bottle is besides a sink in the kitchen''). 
The goal is to retrieve a target image $I_t$ from a candidate pool that contains an instance visually and semantically consistent with the query. Formally, the task can be expressed as:
\begin{equation}
  I_t = \arg\max_{I \in \mathcal{C}} \operatorname{sim}(f(I_q, T_q), f(I)),
  \label{eq:definition}
\end{equation}
where $\mathcal{C}$ is the candidate pool of images, $f(\cdot)$ represents the multimodal embedding function that encodes both the visual instance and textual query into a unified representation space, and $\operatorname{sim}(\cdot, \cdot)$ computes the similarity between embeddings.

\label{sec:bench}

In our benchmark, only the cropped region of the target instance, as defined by the detection bounding box, is used as the query. This approach provides users with a more direct way to specify the reference instance, effectively disregarding the surrounding image context.
We only use full images as candidates because candidate images are usually pre-encoded and stored in databases, where the user's query intent is unknown at the time of encoding. 

We consider two query text types as subtasks in IDMR. Each subtask consists of a reference image $I_q$, A query text $T_q$ and a candidate image pool $\mathcal{C} = \{ I_i \}_{i=1}^M$, where only one candidate is the positive target sample $I_t$. 

 {\bf Instance Search.} Users may be interested in a specific instance and want to find the same instance appearing in different contexts or locations. For example, ``Given the [instance name] in the image, find a new image containing the [instance name]''. It evaluates the models' ability to identify and match the same instance across different contexts.

 {\bf Instance-Location Search.} The query text specifies the location or environment where the instance should appear (e.g., ``Find the bottle placed in the door compartment of a refrigerator.''). 
This task simultaneously evaluates the model's capabilities in instance recognition, textual instruction understanding, and contextual scene understanding.

\subsection{Benchmark Data Construction}

To evaluate generalization, we choose a real-world Object Tracking dataset to construct benchmark data because it ensures continuous visual recordings of a certain object over time. This makes it an ideal resource for assessing the model's ability to retrieve the same instance consistently. We also selected a first-person action dataset because it features the same instances appearing in different positions and contexts, which aligns well with our task definition. 

{\bf LaSOT} \cite{lasot} is a Large-scale Single Object Tracking dataset, which provides continuous visual recordings of a target object over time, making it an ideal resource for evaluating the model's ability to retrieve the same instance.  
We utilize the test set of LaSOT, which contains single-object tracking data across 70 categories, with 4 distinct objects per category. 
To ensure diversity in object backgrounds, we uniformly sample 5 images per object at equal intervals. 
For each instance, we randomly pair queries and positives, constructing the tasks described in \cref{sec:bench}. 
In the Instance Search subtask, the query text is formulated as ``Find the [classname]'', while for Instance-Location Search, the query text is generated using a Multimodal Large Language Model (MLLM), following the same procedure as Step ``MLLM Reasoning'' in the training sample generation process.  

{\bf EPIC-KITCHEN-100} \cite{epic_kitchen} is a large dataset in first-person  vision. 
It has multi-faceted, audio-visual, non-scripted recordings in the native environment, i.e., the wearers' homes, capturing all daily activities in the kitchen over multiple days. 
We manually select 100 image pairs containing the same instance with different image contexts.
The generation of query text also follows that in \cref{fig:data pipeline} (bottom).

We use all the images as candidates in the Location-conditioned Instance Search task. 
We just select 20 images as candidates for the Instance Search task since only the different instances of the same class are hard negative samples. 
Visualized examples and statistics of our test data are provided in Appendix B.

\section{A Cross-domain Training Method}

\subsection{Preliminaries}
In our training setup, each sample consists of a query $q=(I_q, q_t)$ and a positive candidate $c^+=(I_t, c_t^+)$. 
Both the query and the candidate can be text, image or the combination in multimodal embedding tasks. 
The negative samples for a query are randomly selected from the positive candidates of other queries within a batch. 
Since the query and its positive candidate are precisely matched, using random negative samples is sufficient to enable the model to learn a discriminative embedding space. 

To get the final query text $q_t$, an instruction is concatenated to the training query text $T_q$ :
\begin{equation}
  q_t = \{ \text{IMAGE\_TOKEN} \} \{\text{Instruction} \}\{ T_q \},
  \label{eq:query}
\end{equation}
Where the IMAGE\_TOKEN is needed if the query image exists. 
Since our constructed data does not include candidate text, the candidate text $c_t^+$ is defined as follows:
\begin{equation}
  c_t = \{ \text{IMAGE\_TOKEN} \} \{\text{Instruction} \},
  \label{eq:query}
\end{equation}
The implementation of instructions is shown in Appendix A.

\subsection{Training Data Construction}
\begin{table}[h]
\centering
\resizebox{0.99\linewidth}{!}{
\begin{tabular}{@{}lccc@{}}
\toprule
\textbf{Data Source} & \# \textbf{Categories} & \textbf{\# Origin objects} & \textbf{\# Train data} \\
\midrule
COCO               & 80 & 860K & 27K   \\ 
Objects365         & 365& 9.6M & 219K   \\ 
Open images        & 600& 14.6M & 311K   \\ 
\bottomrule
\end{tabular}
}
\caption{Statistics of the training data.}
\label{tab:train_stats}
\end{table}

Public datasets directly relevant to our task are exceedingly rare. The few available single-object tracking datasets serve only as testing benchmarks and cannot be scaled effectively. Therefore, we have explored methods to synthesize large-scale data using more common object detection datasets as our foundation. This approach not only overcomes dataset limitations but also effectively validates our model's ability to generalize across different domains.

{\bf Data Sources.} We construct training data based on three large scale object detection datasets: COCO \cite{coco}, Objects365 \cite{objects365} and Open images \cite{openimages}. 
We hope the model can learn to recognize the same objects (instance-driven abilities) by studying the relationship between cropped images of objects and the corresponding original full images. 
This method has the potential to be scaled up by increasing object detection data. 
We do not construct training data using the same way as constructing evaluation data since it is challenging to mine large scale image pairs with the same instance. 
Note that all training data are collected from the original training set of these above object detection datasets, while we retain the validation dataset as our in-domain test dataset.

{\bf Filter.}
The original dataset contains many small, insignificant annotated objects that may not be visually prominent or relevant. To avoid confusing the MLLM (Qwen2VL-72B-Instruct), we designed a data filtering strategy. Specifically, we consider objects that can be recognized by the CLIP model as suitable training instances. 
As illustrated in \cref{fig:data pipeline} (left), We crop the image based on the annotated bounding box and calculate the similarity score between the cropped object and the text prompt ``a photo of a [classname]'' using the CLIP \cite{clip} model. To filter out hard-to-recognize objects, we set a similarity score threshold of 0.2. Instances with scores below this threshold are excluded, ensuring that only visually meaningful and recognizable objects are retained for training. Moreover, to ensure class balance in our training data, for categories with more than 1000 samples, we only retain 1000 samples.

\begin{figure}
  \centering
    \includegraphics[width=1\linewidth]{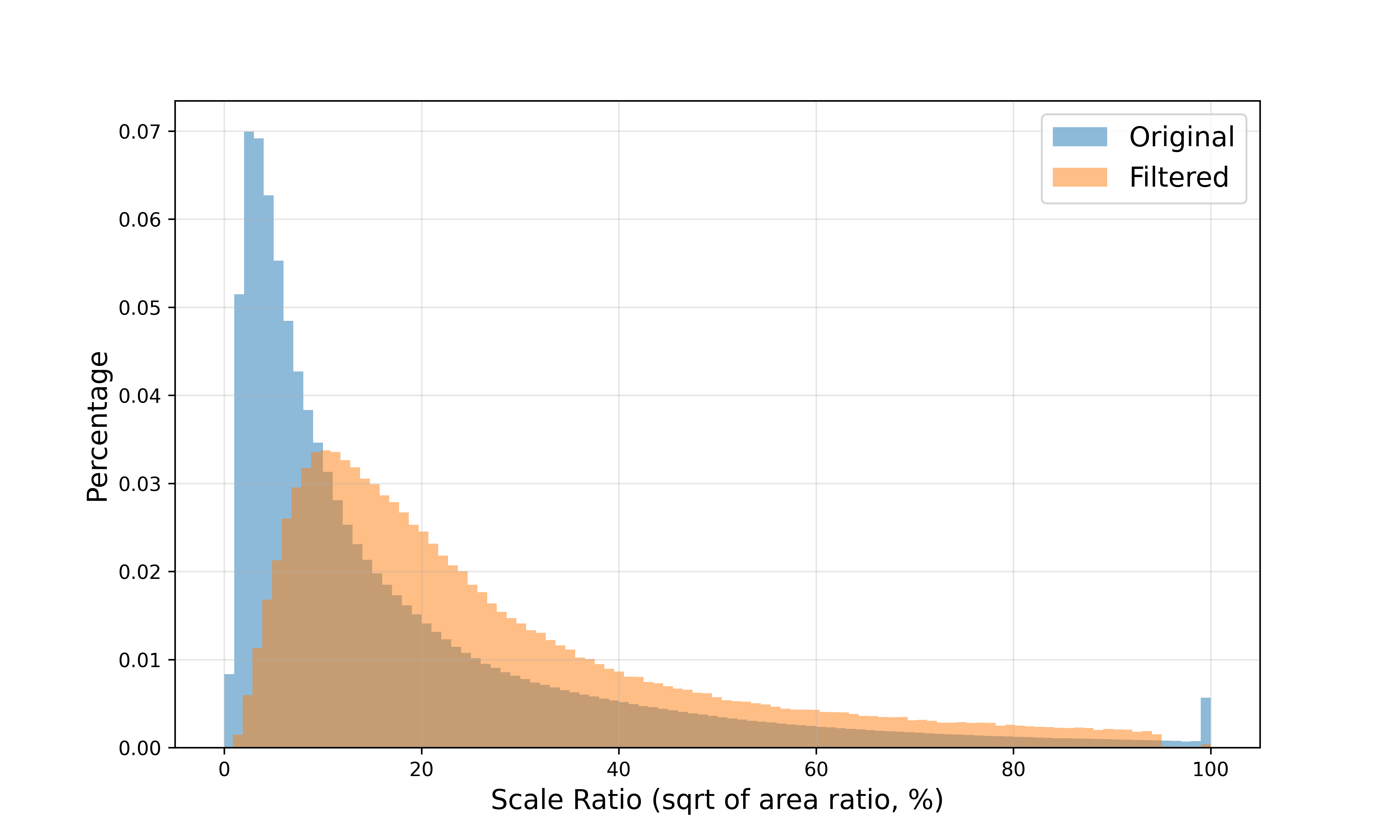}
  \caption{Distributions of bounding boxes in the IDMR training dataset before and after filtering.}
  \vspace{-3mm}
  \label{fig:filter data distribution}
\end{figure}

We show the distribution of the area ratios of cropped images to their original images, as illustrated in \cref{fig:filter data distribution}
The filtering strategy removes objects of very small sizes, which are difficult to recognize and less significant for retrieval purposes. 

\label{set:Generate Query Text.}
{\bf Generate Query Text.}
As shown in \cref{fig:data pipeline} (middle), we utilize open-source MLLM as we used in the filtering process to generate a location-oriented caption of the object. We prompt the MLLM with a visual marker on the input image, which is a green bounding box that frames the target. 
The generated caption is used to build query text in our training data.

{\bf Training Data.}
We construct the image-text to image training triplet $(I_r, T_q, I_c^+)$ with the generated location-oriented caption and the existing annotated bounding box. 
As shown in \cref{fig:data pipeline} (right), we use the cropped object as query image $I_r$, the whole image as the positive candidate image $I_c^+$.
 We show statistics of the training data in \cref{tab:train_stats}. A dataset of 557K triplets is created based on the above data synthesis pipeline.

\subsection{Contrastive Training}

We employ contrastive training to train an embedding model, utilizing the hidden state $\mathbf{h}$ of the last token as the multimodal embedding. The loss function is Noise Contrastive Estimation(InfoNCE) Loss \cite{oord2018representation}, defined as:
\begin{equation}
  \mathcal{L} = -\log \frac{\exp(\mathbf{h}_{q_i} \cdot \mathbf{h}_{c_i^+} / \tau)}{\exp(\mathbf{h}_{q_i} \cdot \mathbf{h}_{c_i^+} / \tau) +
  \sum_{c_j^- \in C^-} \exp(\mathbf{h}_{q_i} \cdot \mathbf{h}_{c_j^-} / \tau)}
  \label{eq:loss}
\end{equation}
where $q_i$ and $c_i^+$ is the $i$-th query and its corresponding positive candidate, while $c_j^-$ denotes the $j$-th negative candidate in a batch. The temperature $\tau$ is a hyper-parameter that controls the concentration of the distribution. 

In contrastive learning, sufficient negative samples are crucial, implying the need for a large batch size. Hence, the GradeCache \cite{gradecache} strategy is adopted during training. It is a gradient caching technique that divides a large batch of data into smaller batches, computes the gradients separately, and accumulates gradients to ensure a large batch size. 

\section{Experiment}

\begin{table*}[h]
\begin{small}
\centering
\begin{tabular}{@{}lc|cccc|cccc|c@{}}
\toprule
\multirow{3}{*}{Model}&\multirow{3}{*}{\# Params} & \multicolumn{4}{c|}{LaSOT} & \multicolumn{4}{c|}{EPIC-KITCHENS-100} & \multirow{2}{*}{Avg} \\
\cmidrule(lr){3-6} \cmidrule(lr){7-10}
& &\multicolumn{2}{c}{Instance} &  \multicolumn{2}{c|}{Location} &  \multicolumn{2}{c}{Instance} &  \multicolumn{2}{c|}{Location} &  \\
& & P@1 & R@5 & P@1 & R@5 & P@1 & R@5 & P@1 & R@5 & P@1 \\ 
\midrule
CLIP           &428M &32   &53.9 & 28.9  &43.6  &15.83  &32.5 &24.2 &46.7 &25.2\\
SigLIP          &878M &38.6 &57.4   & 45.6 &62.1 &19.2  &\underline{45}  &43.3  &72.5 &36.7\\
OpenCLIP        &2.54B &40   &60.1   & 48.3  &69.7 &\underline{20.8} &41.7   &35.8   &61.7  &36.2\\
Magiclens       &465M &21.5     &41.2   & 46.9  &68.9   &9.2   &25   &20.8  &51.7 &24.6\\
VLM2Vec         &4.15B &31.9 &53.6   & 67.3  &84.5   & 8.3  &24.2  &30.8 &70 &34.6 \\
MM-Embed        &7.57B & 39.5  &60.1   &74.2   &88.8   &12.5   &40.8   &47.5   &79.2 &43.4\\
\midrule
Ours          &8.1B &\underline{46.7}  &\underline{67.8}   &\underline{88.1}  &\textbf{94.9}   &17.5   &\underline{45}  &\underline{59.2} &\textbf{86.7}  &{52.9} \\
    \rowcolor{gray!25}
Ours        &25.57B &\textbf{59.7}  &\textbf{76.4}   & \textbf{88.8}  &\underline{94.6}   &\textbf{21.7}   &\textbf{52.5}  &\textbf{60} &\underline{85.8}  &\textbf{57.6}\\
\bottomrule
\end{tabular}
\vspace{-2mm}
\caption{{\bf Zero-shot} evaluations across different models on IDMR-bench. Precision@1 and Recall@5 scores are provided for each meta-task as well as the average Precision@1 scores.}
\vspace{-4mm}
\label{tab:performance-mybench}
\end{small}
\end{table*}

\subsection{Settings}
{\bf Configuration.}
We train our model using 1.2 million interleaved text-image pairs. Of these, 557K pairs are synthetic data constructed by our proposed method. To enhance the model's basic retrieval capabilities, we leverage the publicly available MMEB training set, which provides an additional 662K multimodal pairs.
We adopt the pretrained InternVL2.5 \cite{internvl2.5} as our backbone model. For parameter-efficient fine-tuning, we employ Low-Rank Adaptation (LoRA) \cite{hu2022lora} with a rank of 8. The training is conducted with a batch size of 256 and a learning rate of 2e-5 with linear decay. All experiments are trained for a single epoch.

\noindent \textbf{Benchmarks.}
We conduct extensive experiments on three kinds of benchmarks. Firstly, we randomly synthesize 1K data samples from the validation dataset of COCO \cite{coco}, Objects365 \cite{objects365} and Open images \cite{openimages}, using the same way of our training data construction pipeline. We name it as in-domain test data. The second benchmark is our processed zero-shot evaluation datasets to assess the model's generalization capabilities.

To evaluate the multimodal retrieval capabilities of our model, we also use the MMEB \cite{vlm2vec}, which is a comprehensive benchmark that integrates four tasks: classification, visual question answering (VQA), retrieval, and visual grounding. 
This benchmark is specifically designed to validate the embedding capabilities of multimodal models across various combinations of image-text data.
Unless otherwise specified, we adopt Precision@1 as the primary evaluation metric across all experiments.

\noindent \textbf {Baseline.}
We compare our model against several strong baselines, ranging from foundational vision-language models to state-of-the-art MLLM-based multimodal embedding approaches. These include CLIP \cite{clip}, SigLIP \cite{siglip}, OpenCLIP \cite{openclip}, MagicLens \cite{magiclens}, VLM2Vec \cite{vlm2vec}, and MM-Embed \cite{lin2024mm-embed}.

\subsection{Results}

\subsubsection{Performance on In-domain Test Dataset}

The Precision@1 performance on in-domain test data is shown in Table \ref{tab:performance-mybench-indoman}. Our model significantly outperforms all baselines across all three test sets. 
Compared to the strongest baseline MM-Embed \cite{lin2024mm-embed}, our 8B model demonstrating a substantial average improvements of 34.1\% with a similar model size. This confirms that our approach effectively captures fine-grained visual correspondence between object instances.

\begin{table}[h]
\begin{small}
\centering
\begin{tabular}{@{}l|c c c@{}}
\toprule
\multirow{2}{*}{Model} & \multicolumn{3}{c}{In-domain Test Data} \\
\cmidrule(lr){2-4}
& COCO & Objects365  & Open Images \\
\midrule
CLIP            &22.3   &13.8   & 38.4  \\
SigLIP          &34.3   &20.2   & 48.9   \\
OpenCLIP        &37.9   &23.7   & 50.3     \\
Magiclens       &24   &12   & 22.7     \\
VLM2Vec         &45.2   &27   & 52.5     \\
MM-Embed        &44.9   &27.3 & 54.5      \\
\midrule
Ours (8B)        &\underline{74.8} &\underline{69.6}   &\underline{84.7}    \\
    \rowcolor{gray!25}
Ours (26B)       &\textbf{78}   &\textbf{78.4}  &\textbf{91.7}    \\
\bottomrule
\end{tabular}
\vspace{-2mm}
\caption{Performance comparison across different models on the in-domain test dataset.}
\vspace{-4mm}
\label{tab:performance-mybench-indoman}
\end{small}
\end{table}

\subsubsection{Performance on Zero-shot IDMR-bench}

To rigorously assess the model's generalization capabilities in real-world scenarios, we evaluate it on zero-shot tasks constructed from LaSOT and EPIC-KITCHENS-100. 
We construct two tasks on each dataset. Each task have different prompt instructions. The details of zero-shot evaluation implementations can be found in Appendix A.

\begin{table*}[ht]
\begin{small}
\centering
\begin{tabular}{@{}lc|cccc|ccc@{}}
\toprule
\multirow{2}{*}{Models} &\multirow{2}{*}{\# Params} & \multicolumn{4}{c|}{Per Meta-Task Score} & \multicolumn{3}{c}{Average Score} \\
\cmidrule(lr){3-6} \cmidrule(lr){7-9}
&& Class. & VQA & Retr. & Ground. & IND & OOD & Overall \\
\midrule
\# datasets&- & 10 & 10 & 12 & 4 & 20 & 16 & 36 \\
\midrule
CLIP &428M & 55.2 & 19.7 & 53.2 & 62.2 & 47.6 & 42.8 & 45.4 \\
OpenCLIP &2.54B & 56.0 & 21.9 & 55.4 & 64.1 & 50.5 & 43.1 & 47.2\\
VLM2Vec (LLaVA-1.6) &7.57B& 54.7 & 50.3 & 56.2 & 64.0 & 61.0 & 47.5 & 55.0 \\
VLM2Vec (Phi-3.5-V) &4.15B& 54.8 & 54.9 & 62.3 & 79.5 & 66.5 & 52.0 & 60.1 \\
MMRet (LLaVA-1.6) &7.57B& 56.0 & 57.4 & 69.9 & 83.6 & 68.0 &\underline{59.1} & 64.1 \\
\midrule
Ours (8B)    &8.1B &\underline{58.3}  &\underline{58.6} &\underline{68.7} &\underline{85.6} &\underline{70.5} &57.9 &\underline{64.9}  \\
    \rowcolor{gray!25} 
Ours (26B)    &25.57B&\textbf{66.3} &\textbf{61.9} &\textbf{71.1} &\textbf{88.6} &\textbf{73.4} &\textbf{63.9} &\textbf{69.2}   \\
\bottomrule
\end{tabular}
\vspace{-2mm}
\caption{Performance comparison across different models on {\bf MMEB} \cite{vlm2vec}. Scores are provided for each meta-task as well as the average score. All models listed in the table have been fine-tuned for 1 epoch using the 662K MMEB training data. In contrast, our models have been just trained for 1 epoch utilizing both the 662K MMEB training data and 557K synthetic data samples.}
\vspace{-4mm}
\label{tab:performance-mmeb}
\end{small}
\end{table*}

\begin{figure*}[t]
  \centering
  \begin{subfigure}[b]{0.46\linewidth}
    \centering
    \includegraphics[width=\linewidth]{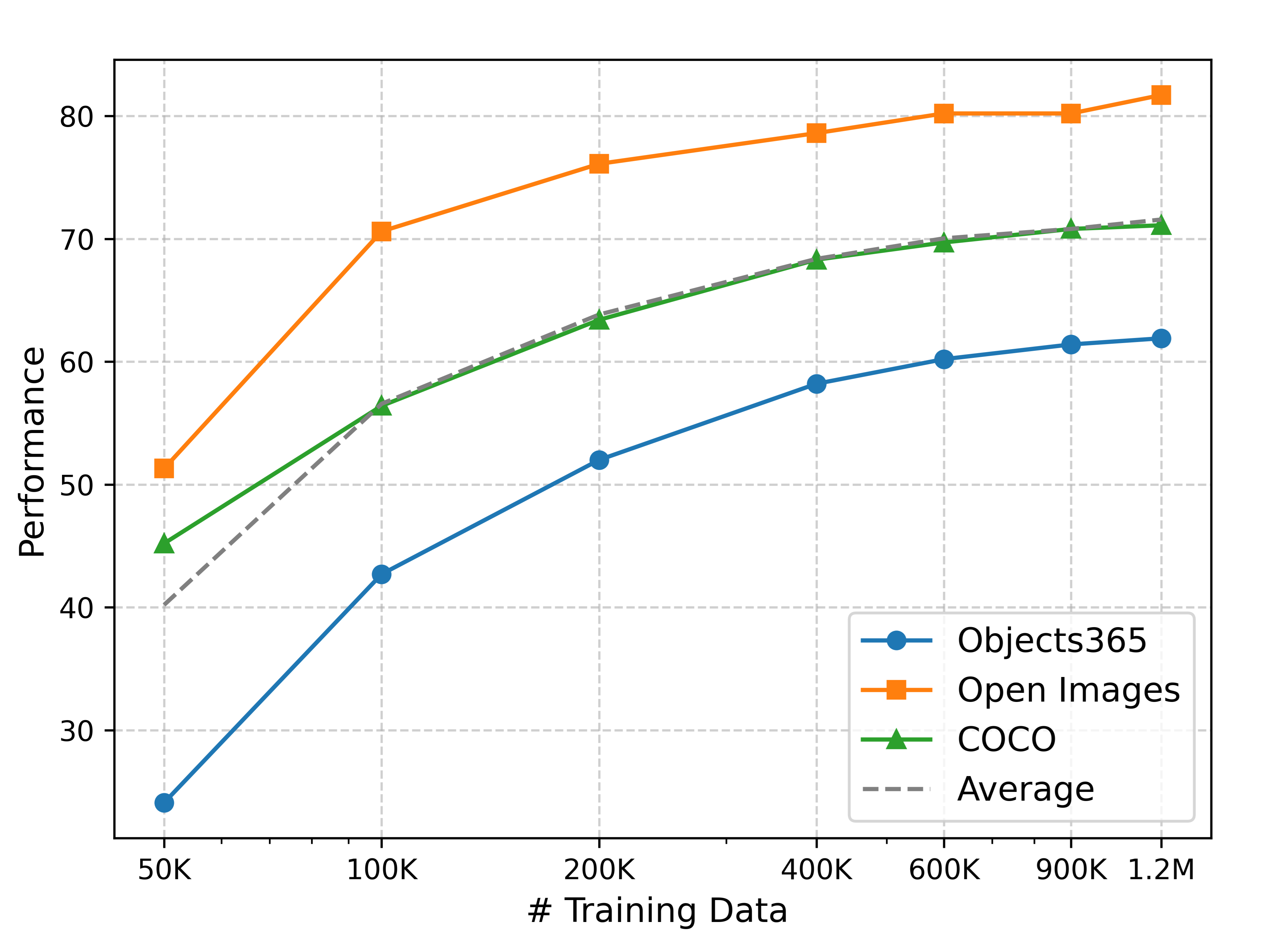}
    \label{fig:data scale}
  \end{subfigure}
  \hfill
  \begin{subfigure}[b]{0.46\linewidth}
    \centering
    \includegraphics[width=\linewidth]{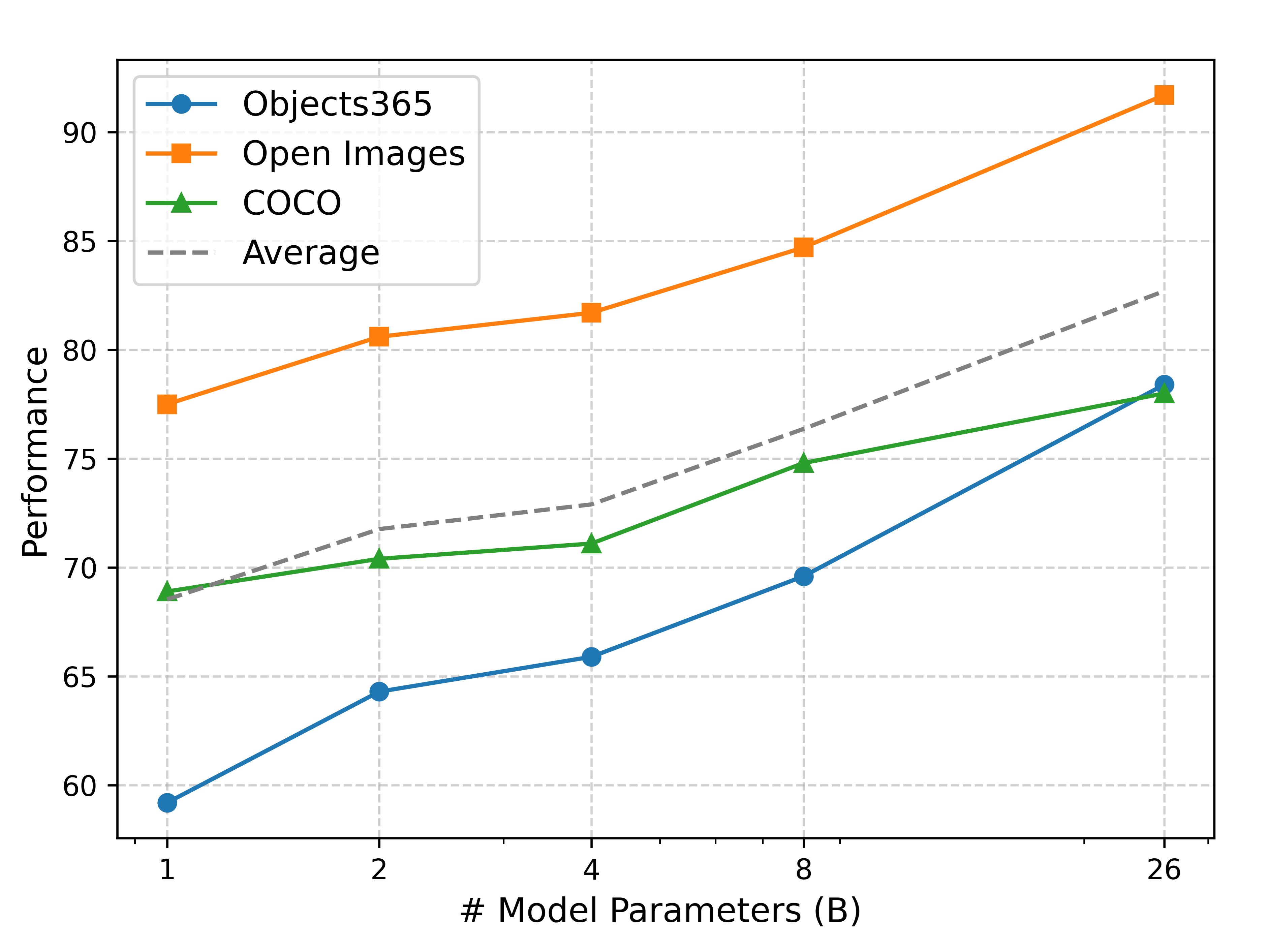}
    \label{fig:model size scale}
  \end{subfigure}
  \vspace{-4mm}
  \caption{Scaling experiments: (a) left: performance with different numbers of training data and (b) right: performance with different sizes of InternVL2.5.}
  \label{fig:scaling}
  \vspace{-4mm}
\end{figure*}

As detailed in Table \ref{tab:performance-mybench}, our model demonstrates superior performance across all subtasks. For the LaSOT benchmark, our 26B model achieves 59.7\% and 88.8\% Precision@1 on Instance Search and Instance-Location Search task respectively, outperforming the strongest baseline by 19.7\% and 14.6\%. Similarly, on the more challenging EPIC-KITCHENS-100 benchmark, our model achieves 21.7\% and 60.0\% Precision@1, surpassing the best baseline by 0.9\% and 12.5\% respectively.

It is noteworthy that, despite the Instance Search task being an unseen task for our model, it achieves state-of-the-art results.
These results demonstrate that our training methodology enables effective generalization to unseen domains and tasks, confirming the robustness of our approach for instance multimodal retrieval in real-world applications.

\subsubsection{Performance on MMEB benchmark}

As shown in Table~\ref{tab:performance-mmeb}, our model based on InternVL2.5-26B achieves state-of-the-art performance across all meta-task categories, with an overall average Precision@1 of 69.2\%. 
These results represent significant improvements over previous state-of-the-art models, with gains of 8.03\%, 3.32\%, 1.21\%, and 3.0\% compared to MMRet, the previous best-performing model.
It is worth highlighting that our model achieves these results after being trained for just one epoch on a combined dataset of 662K MMEB training samples and 557K synthetic data samples generated by our method.

Our 8B parameter model also outperforms previous state-of-the-art models, achieving an overall precision@1 of 64.9\% across all 36 datasets. 
Notably, it performs better than MMRet, even though MMRet leverages 26M data points with a model of similar size.
This indicates that the improvements from our approach are not solely attributable to model scale, but rather to the effectiveness of our instance-driven training approach and data construction pipeline.

These results collectively demonstrate that our instance-driven training approach not only excels at the specific task of instance-driven multimodal retrieval but also enhances general multimodal embedding capabilities across diverse tasks and domains. 
This suggests that fine-grained visual correspondence is a fundamental capability that benefits a wide range of multimodal understanding tasks. 

\subsection{Discussion and Analysis}

\subsubsection{Scaling}

{\bf Scaling of Training Data Size.}

Since our training data construction pipeline can easily scale to a large volume, it is essential to analyze whether the model benefits from the increased amount of training data. Models are trained on InternVL2.5-4B for 1 epoch, batchsize 256 and LORA rank 8.
As shown in \cref{fig:scaling} (a), the performance of our in-domain test dataset consistently improves when data size increases from 50K to 1.2M. 

This scaling pattern aligns with observations in other multimodal embedding tasks \cite{vlm2vec, gme}, suggesting that our synthetic data construction method is effective. 
These results highlight the importance of data diversity and volume in training instance-driven multimodal retrieval models, while also indicating that our approach can achieve strong performance with moderate data sizes.

{\bf Scaling of Model Size.}
We investigate how model size affects instance-driven multimodal retrieval performance using various InternVL2.5 models with parameters ranging from 1B to 26B. Models are trained with 1.2M training data for 1 epoch, batchsize 256 and LORA rank 8. 
As shown in \cref{fig:scaling} (b), performance on our in-domain test dataset exhibits linear improvements as model size increases.
This demonstrates that our training data effectively scales with model capacity and larger models consistently improve instance-driven retrieval accuracy without saturation. 

\subsubsection{Hyperparameters}
To evaluate the generalizability of our approach, we conducted a series of hyperparameter sensitivity experiments to investigate how different configurations impact model performance. Table \ref{tab:hyperparameters} presents the experimental results under various training settings.

\noindent\textbf{Impact of LoRA Rank.} We first studied the effect of LoRA adapter rank on the performance of the 4B model. Experiments show that a smaller LoRA rank (rank=4) yields optimal performance across all evaluation metrics compared to higher rank configurations.

\noindent\textbf{Impact of Batch Size.} 
We analyze batch size effects (per GPU) on our 8B model using 8 GPUs (effective batch size = 8×reported). Surprisingly, smaller batches (128 per GPU, 1024 total) achieve peak accuracy (65.4\%), while larger batches (1024 per GPU, 8192 total) degrade performance by 2\%. This contrasts with standard contrastive learning where larger batches typically help by providing more negatives. We hypothesize that: (1) moderate gradient noise from smaller batches prevents overfitting to instance-specific features, and (2) extremely large batches may oversmooth the loss landscape for fine-grained retrieval tasks.

\begin{table}[h]
\centering
\begin{small}
\begin{tabular}{l|c c c c}
\toprule
Setting & In-domain & Zero-shot  & Average \\
\midrule
\multicolumn{4}{c}{Fine-tuning LORA rank (4B)} \\
\midrule
rank=4 & 74.7  &50.4    &62.6  \\
rank=8 & 71.6 &45.8   & 58.7 \\
rank=16 & 64.3 &46.8   &55.6  \\
\midrule
\multicolumn{4}{c}{Training batchsize per GPU (8B)} \\
\midrule
batchsize=128 &76.5  &54.3    &65.4  \\
batchsize=256 &76.4  &52.9    &64.7  \\
batchsize=512 &75.7  & 52.7   &64.2  \\
batchsize=1024 &74  &52.7   &63.4  \\
\bottomrule
\end{tabular}
\caption{Ablation Study of different training settings. }
\label{tab:hyperparameters}
\end{small}
\end{table}

\section{Conclusion}
In this work, we introduce Instance-driven Multimodal Image Retrieval (IDMR), a novel task addressing the scenario of retrieving images containing the same instance based on a query image and descriptive text. Our work makes three key contributions: (1) the creation of IDMR-bench, a comprehensive benchmark designed to evaluate models' ability to retrieve images containing identical instances in contexts matching textual descriptions; (2) a scalable method for automatically constructing large-scale instance-driven image-text to image triplets from object detection datasets; and (3) a powerful multimodal embedding model trained on our synthetic data that achieves state-of-the-art performance on both IDMR-bench and general multimodal embedding tasks. Our experiments reveal the limitations of existing multimodal retrieval methods in performing fine-grained instance-level matching, demonstrating the substantial improvements our approach brings to this challenging task. Due to computational constraints, our current experiments are limited to 26B-parameter models and 1.2M training samples. In future work, we plan to develop and release 10M-scale training datasets and 78B-parameter multimodal retrieval models to further advance research in this important direction.

{
    \small
    \bibliographystyle{plain}
    \bibliography{main}
}


\clearpage
\section*{Appendix}
\appendix
\label{A}
\section{Data Instructions}
For train data, in-domain test data and zero-shot Instance-Location Search task, the query instruction is: ``Find me an image containing the object in the given image with the following caption''. To create the final query text, the provided caption is concatenated with questions generated by the MLLM.
For zero-shot Instance Search task, the query text is: ``Given the [instance name] in the image, find a new image containing the [instance name]''. 
For candidates that only image modality are available, the instruction is: ``Represent the given image''.

\section{More Details of IDMR dataset}
In this section, we provide statistics of our IDMR zero-shot test data in \cref{tab:benchmark_stats}
We also visualize the examples of the synthetic training data (\cref{tab:visualize IDMR training dataset}) and zero-shot test data (\cref{tab:visualize IDMR testing dataset}) for Instance-Drien Multimodal Retrieval.

\section{Qualitative Results of IDMR model}
\begin{figure*}[h!]
  \centering
   \includegraphics[width=0.95\linewidth]{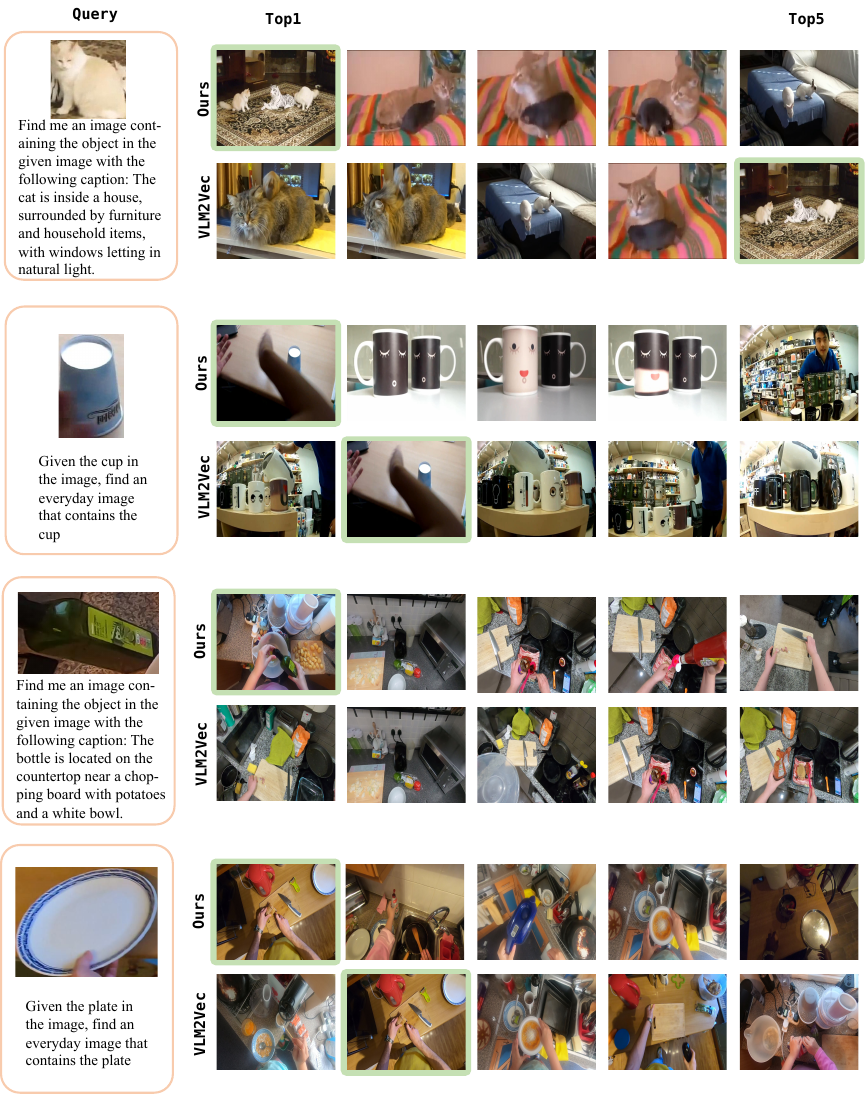}
   \caption{Visualization of the Top-5 results from our model v.s. VLM2Vec on zero-shot test data, with the correct target image highlighted in green.}
   \label{fig:visualize top-5 retrieved images}
   \vspace{-6mm}
\end{figure*}

\begin{figure*}[h!]
  \centering
   \includegraphics[width=0.95\linewidth]{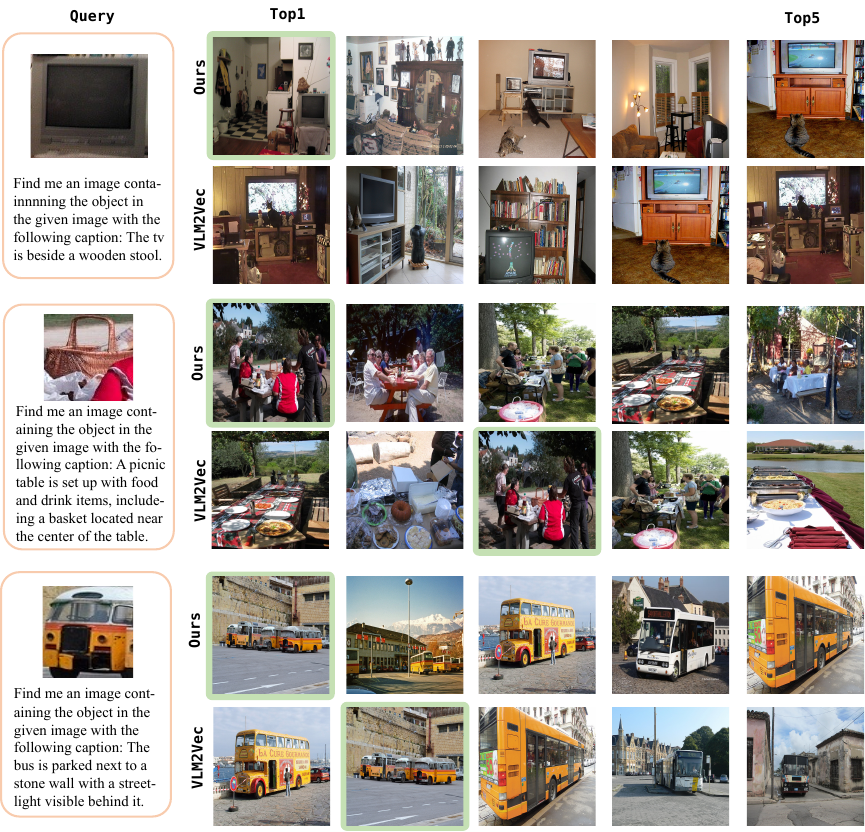}
   \caption{Visualization of the Top-5 results from our model v.s. VLM2Vec on in-domain test data, with the correct target image highlighted in green.}
   \label{fig:visualize in-domain top-5 retrieved images}
   \vspace{-6mm}
\end{figure*}

Top-5 retrieved images of our IDMR model and VLM2Vec \cite{vlm2vec} on IDMR in-domain test set and zero-shot test data are shown in \cref{fig:visualize top-5 retrieved images} and \cref{fig:visualize in-domain top-5 retrieved images} respectively.

\section{Full Results on MMEB}
In \cref{tab:MMEB full results}, we present detailed results on the MMEB benchmark \cite{vlm2vec}. 
This benchmark includes 20 in-distribution datasets and 16 out-of-distribution (OOD) datasets, with the OOD datasets highlighted in blue in the table. 
Our 26B IDMR model achieves state-of-the-art performance across these tasks. 
Additionally, our 8B model outperforms MMRet, even though MMRet has a comparable model size and is trained on 26M data.

\begin{table*}[h]
\centering
\begin{tabular}{lcccc}
\toprule
\textbf{Base Dataset} & \textbf{Qry. Text} & \textbf{Qry. Image} & \textbf{\# Triplets} & \textbf{\# Candidate Images} \\
\midrule
\multirow{4}{*}{LaSOT \cite{lasot}}  & \multirow{2}{*}{Instance Search}    & Crop   & 1400 & 20 \\ 
&                               & Full   & 1400 & 20 \\
& \multirow{2}{*}{Location conditioned Instance Search}    & Crop   & 1400 & 1000 \\ 
&                               & Full   & 1400 & 1000 \\
\cmidrule(lr){1-5}
\multirow{4}{*}{EPIC-KITCHENS-100 \cite{epic_kitchen}}  & \multirow{2}{*}{Instance Search}    & Crop   & 120 & 100 \\ 
&                               & Full   & 120 & 100 \\
& \multirow{2}{*}{Location conditioned Instance Search}    & Crop   & 120 & 100 \\ 
&                               & Full   & 120 & 100 \\
\bottomrule
\end{tabular}
\caption{Statistics of the four zero-shot subsets in IDMR-bench.}
\label{tab:benchmark_stats}
\end{table*}

\begin{table*}[h]
\centering
\begin{tabular}{l|p{3cm} p{6cm} p{3cm}}
\toprule
Dataset & Query Image & Query Text  & Traget Image \\
\midrule
COCO \cite{coco} & \raisebox{-\height}{\includegraphics[width=2cm]{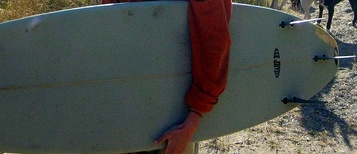}} & Find me an image containing the object in the given image with the following caption: The surfboard is being held by a person wearing an orange shirt and a beanie, with a sandy path and trees in the background.   & \raisebox{-\height}{\includegraphics[width=2cm]{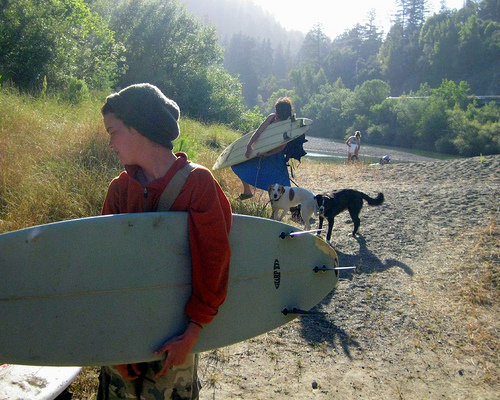}}  \\
\midrule

COCO \cite{coco} & \raisebox{-\height}{\includegraphics[height=2cm]{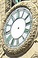}} & Find me an image containing the object in the given image with the following caption: The clock is in the middle of the building.   & \raisebox{-\height}{\includegraphics[height=2cm]{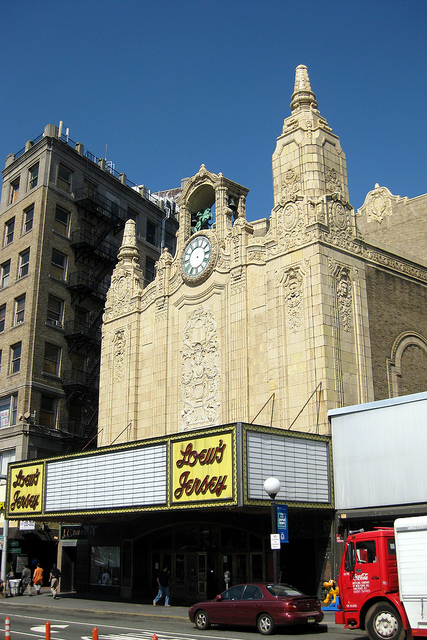}} \\
\midrule

COCO \cite{coco} & \raisebox{-\height}{\includegraphics[width=2cm]{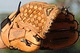}} &  Find me an image containing the object in the given image with the following caption: The baseball glove is located on the pitcher's left hand.   & \raisebox{-\height}{\includegraphics[width=2cm]{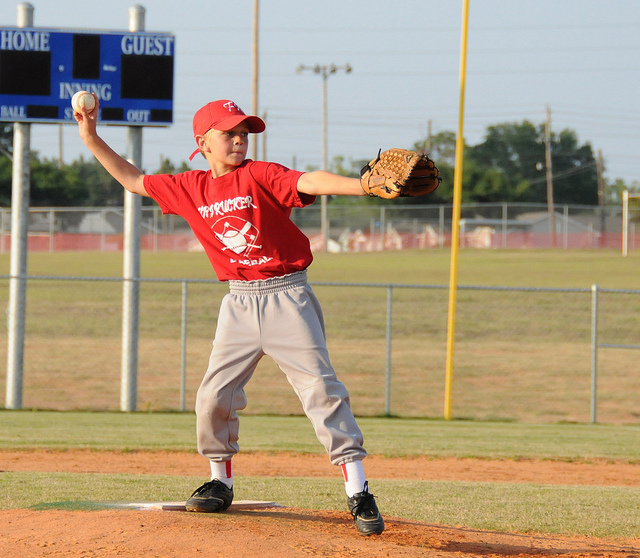}}  \\
\midrule

Objects365 \cite{objects365} & \raisebox{-1.6cm}{\includegraphics[width=2cm]{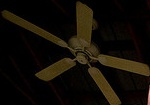}}  & Find me an image containing the object in the given image with the following caption: The fan is located on the ceiling near a person playing guitar.    & \raisebox{-1.6cm}{\includegraphics[width=2cm]{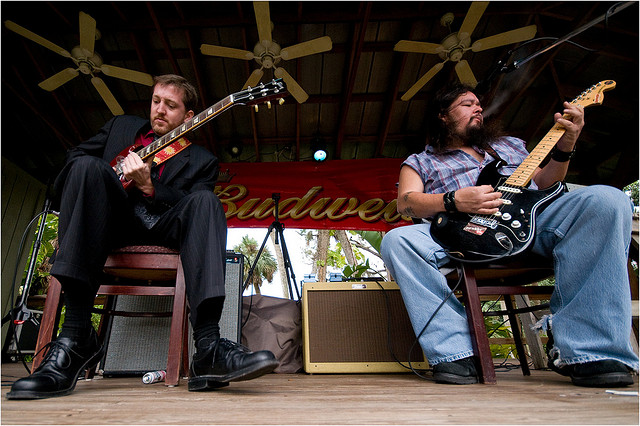}}  \\
\midrule

Objects365 \cite{objects365} & \raisebox{-\height}{\includegraphics[height=2cm]{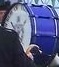}}  & Find me an image containing the object in the given image with the following caption: The drum is located in front of the group of people dressed in kilts.    & \raisebox{-\height}{\includegraphics[width=3cm]{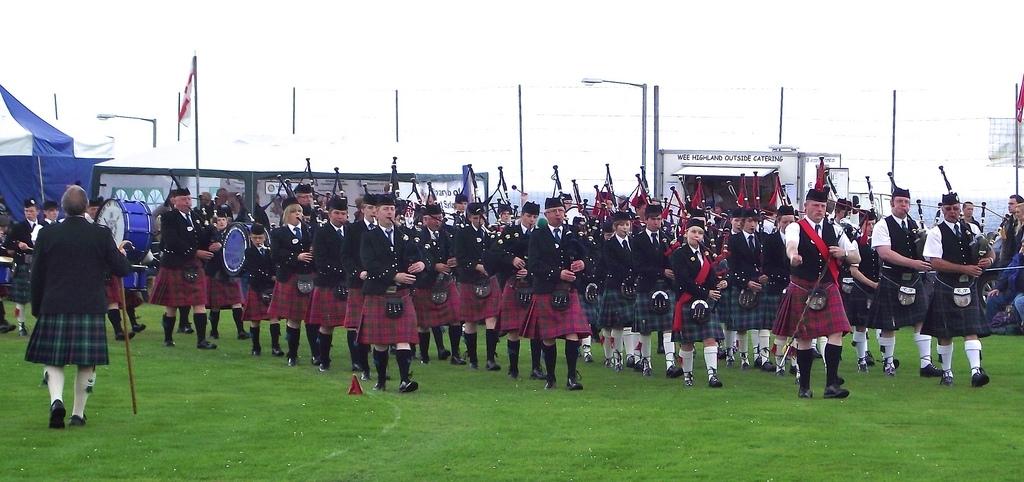}}  \\
\midrule
 
Objects365 \cite{objects365} & \raisebox{-\height}{\includegraphics[width=2cm]{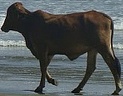}}  &  Find me an image containing the object in the given image with the following caption: The cow is near the shoreline of a beach with waves rolling in.   & \raisebox{-\height}{\includegraphics[width=2cm]{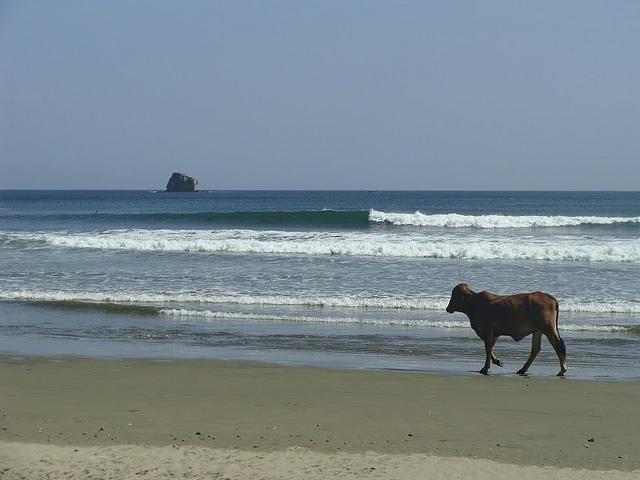}} \\
\midrule

OpenImages \cite{openimages} & \raisebox{-\height}{\includegraphics[width=2cm]{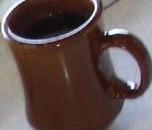}}  &  Find me an image containing the object in the given image with the following caption: The Jug is placed on a table, next to a plate of food and a glass.    & \raisebox{-\height}{\includegraphics[width=2cm]{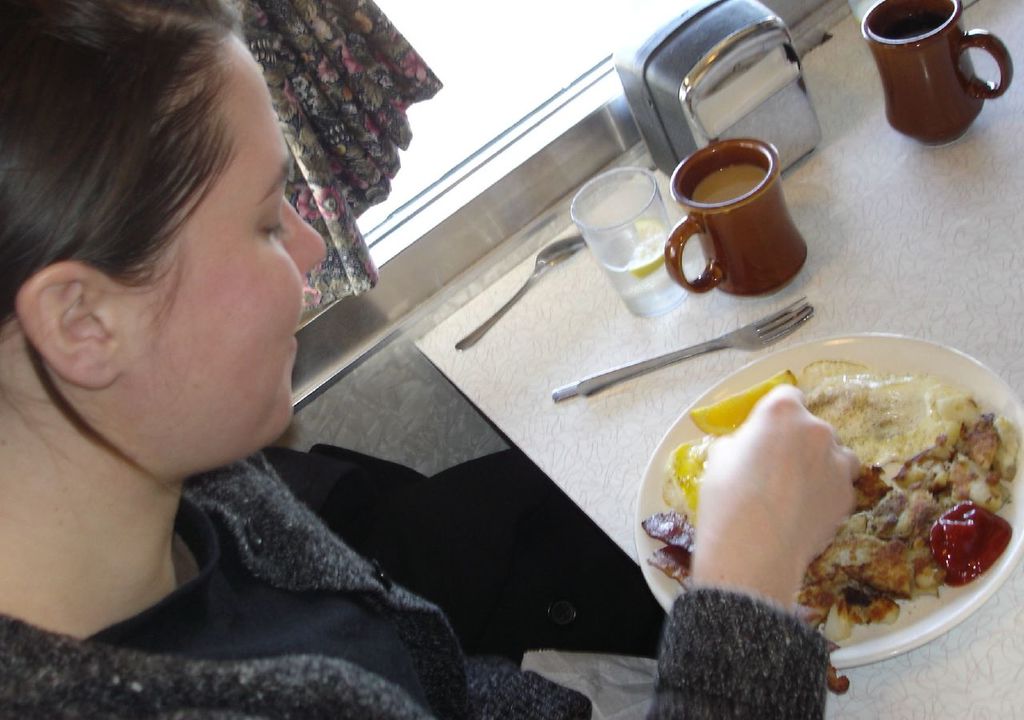}}  \\
 \midrule
OpenImages \cite{openimages} & \raisebox{-\height}{\includegraphics[width=2cm]{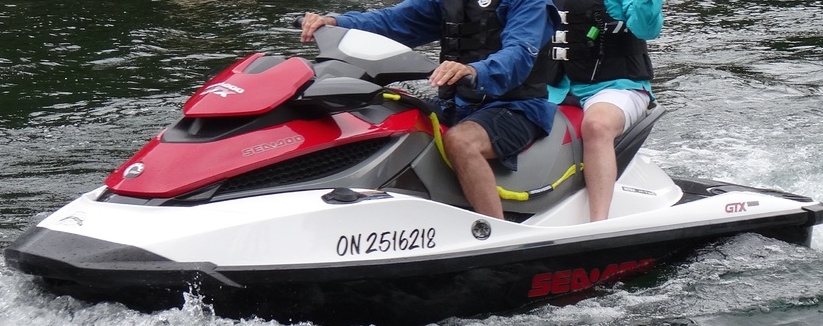}} &  Find me an image containing the object in the given image with the following caption: The Jet ski is on a body of water near a dock and a house with a stone wall.   & \raisebox{-\height}{\includegraphics[width=2cm]{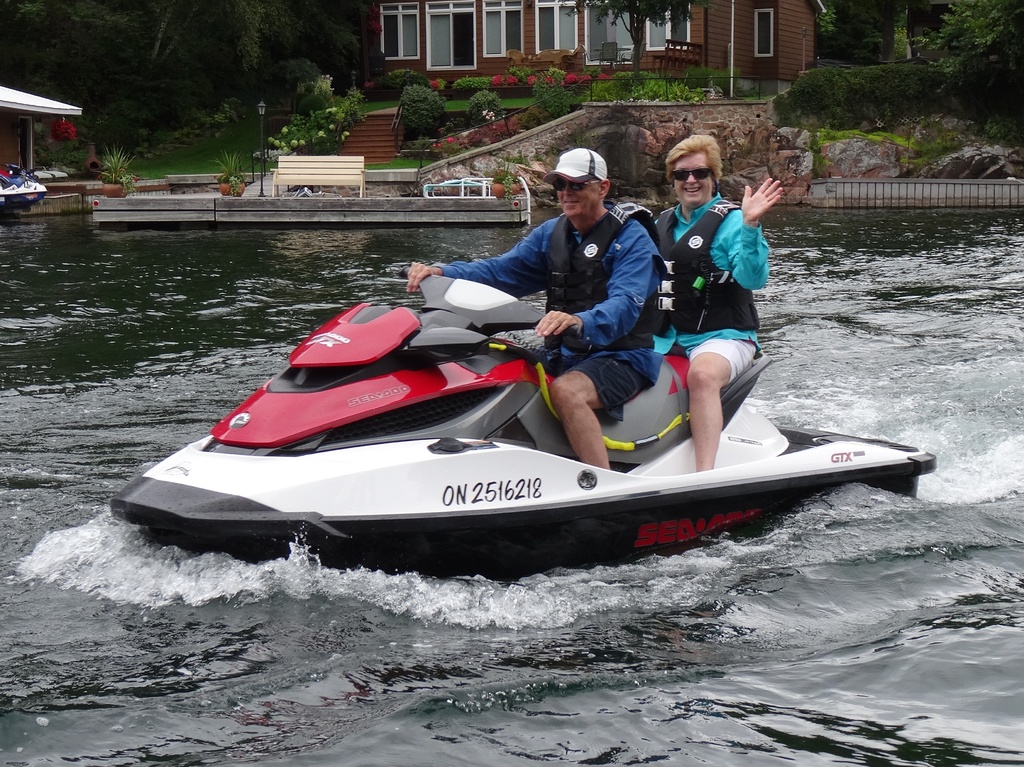}}  \\
 \midrule
OpenImages \cite{openimages} & \raisebox{-\height}{\includegraphics[width=2cm]{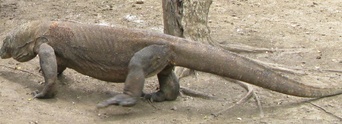}}  &  Find me an image containing the object in the given image with the following caption: The Lizard is situated on a dirt ground near a wooden structure, surrounded by trees and other reptiles.   & \raisebox{-\height}{\includegraphics[width=2cm]{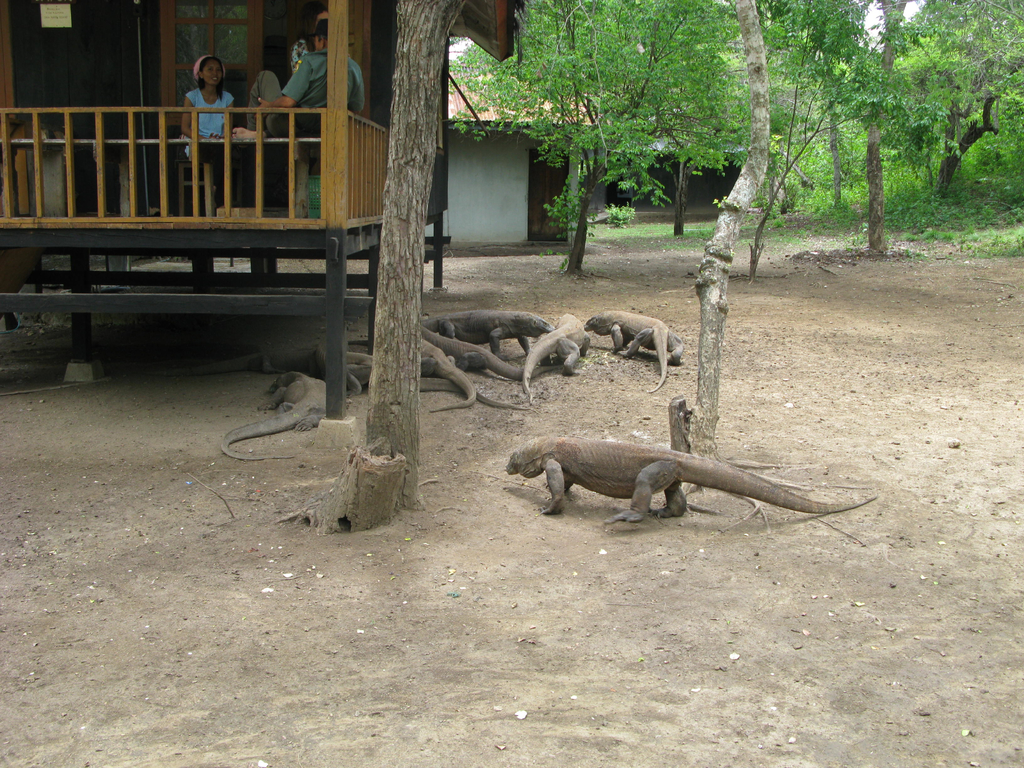}}  \\
\bottomrule
\end{tabular}
\caption{Examples of IDMR training dataset. }
\label{tab:visualize IDMR training dataset}
\end{table*}

\begin{table*}[h]
\centering
\begin{tabular}{l c|p{3cm} p{6cm} p{3cm}}
\toprule
Dataset & Subtask & Query Image & Query Text  & Traget Image \\
\midrule
LaSOT \cite{lasot} & Instance & \raisebox{-\height}{\includegraphics[width=2cm]{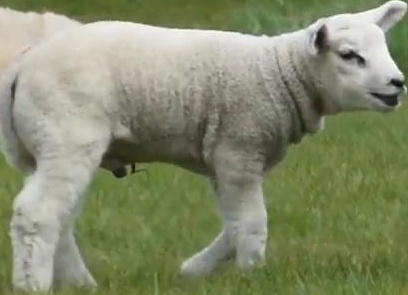}} & Given the sheep in the image, find an everyday image that contains the sheep.   & \raisebox{-\height}{\includegraphics[width=3cm]{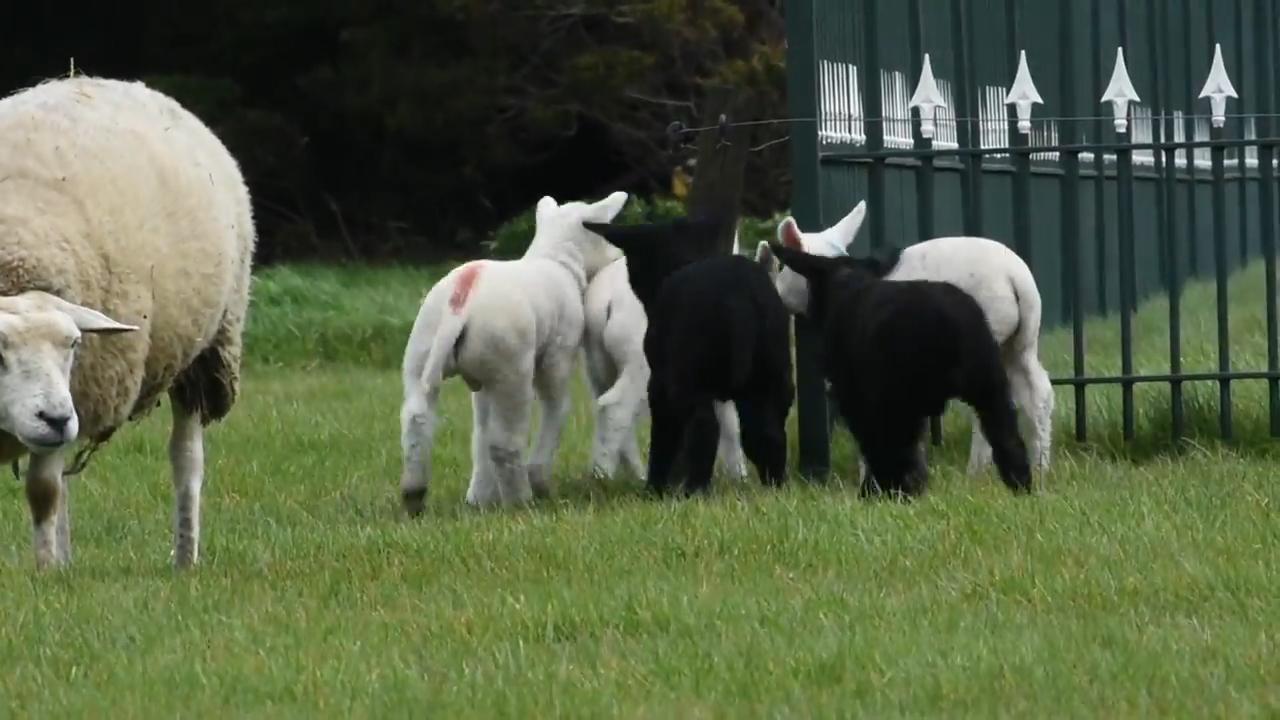}}  \\
\midrule

LaSOT \cite{lasot}& Instance & \raisebox{-\height}{\includegraphics[height=2cm]{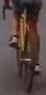}} & Given the bicycle in the image, find an everyday image that contains the bicycle. & \raisebox{-\height}{\includegraphics[width=3cm]{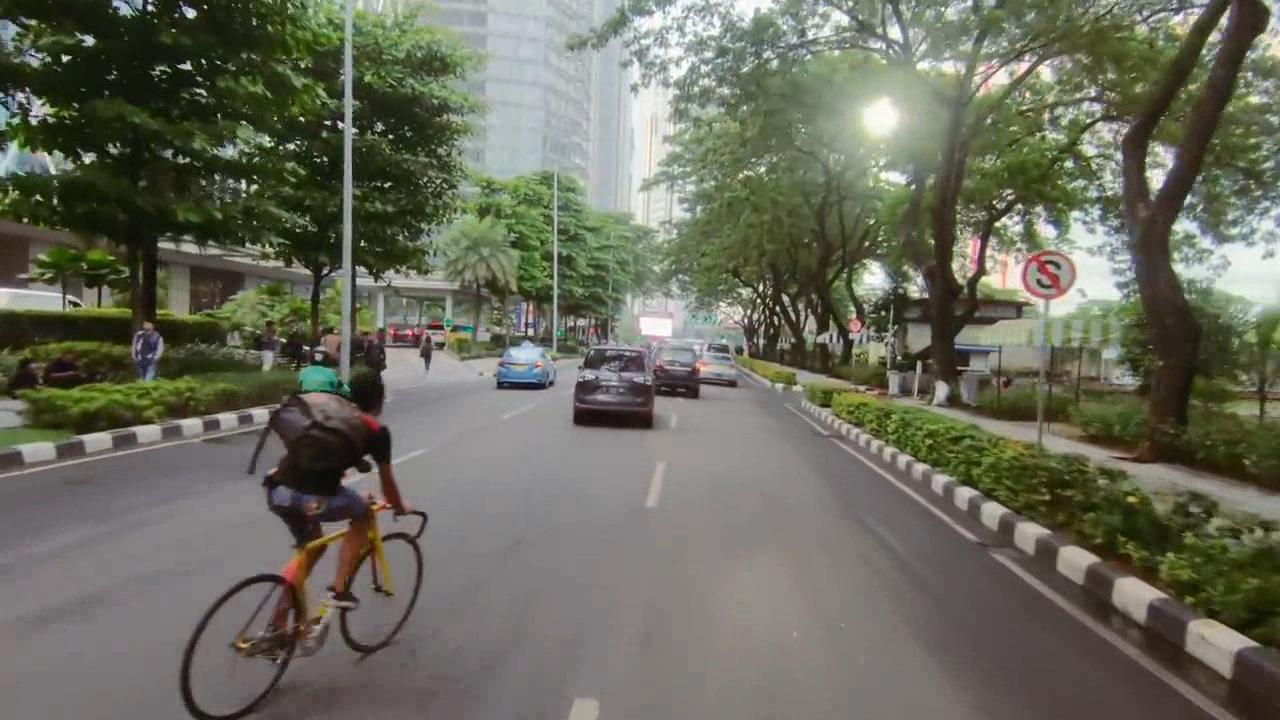}} \\
\midrule

LaSOT \cite{lasot}& Location & \raisebox{-\height}{\includegraphics[width=2cm]{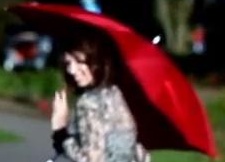}} &   Find me an image containing the object in the given image with the following caption: Find a picture that the umbrella is being held by a woman on a paved path with greenery in the background.   & \raisebox{-\height}{\includegraphics[width=3cm]{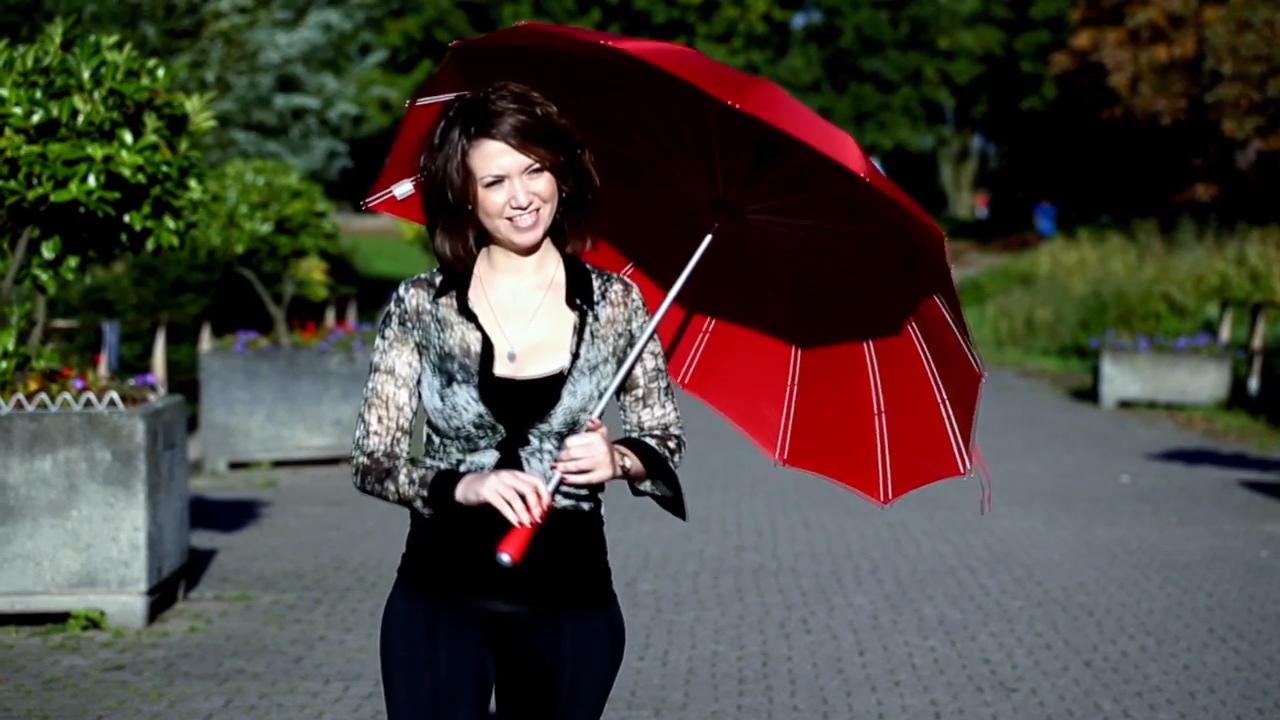}}  \\
\midrule

LaSOT \cite{lasot}& Location & \raisebox{-1.6cm}{\includegraphics[width=2cm]{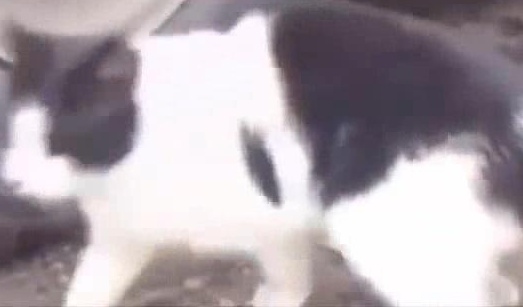}}  &  Find me an image containing the object in the given image with the following caption: The cat is sitting on a patch of dirt or gravel.    & \raisebox{-1.6cm}{\includegraphics[width=3cm]{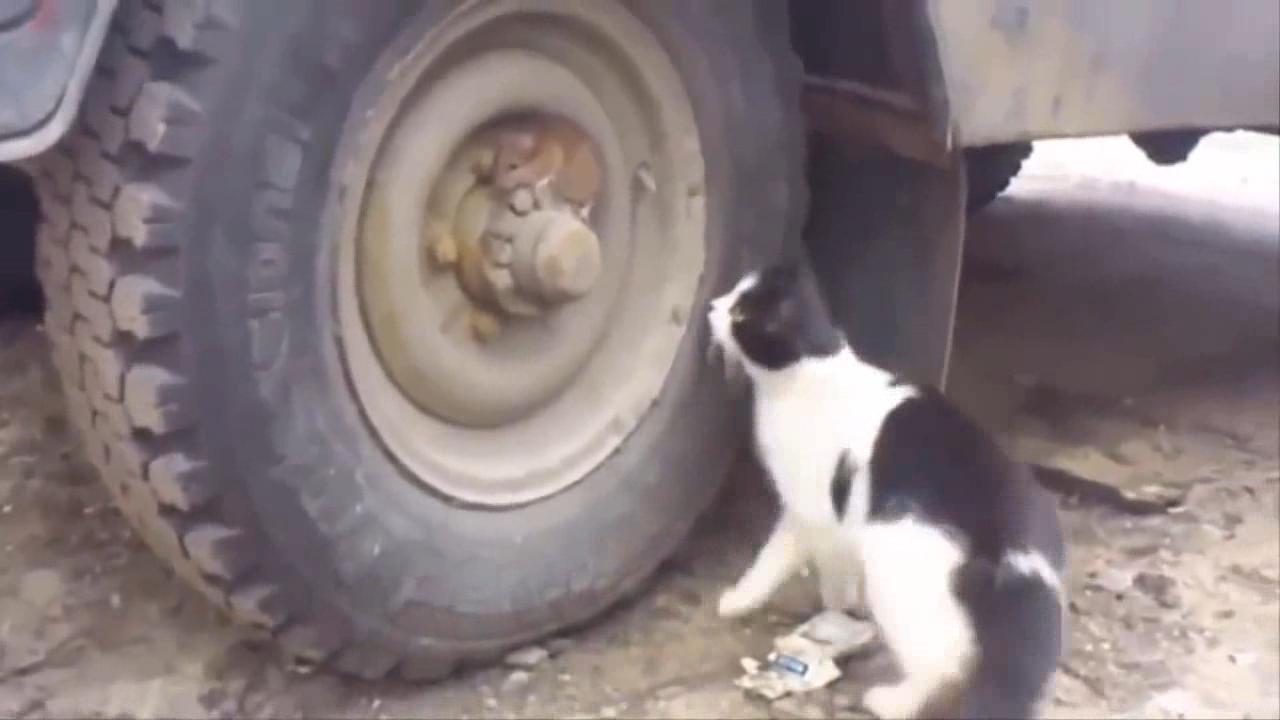}}  \\
\midrule

Kitchens \cite{epic_kitchen} & Instance & \raisebox{-\height}{\includegraphics[width=2cm]{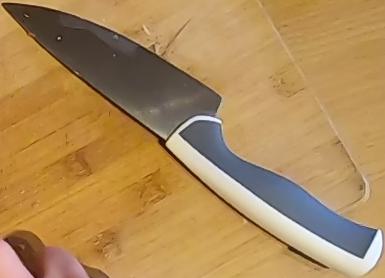}}  & Given the knife in the image, find an everyday image that contains the knife.    & \raisebox{-\height}{\includegraphics[width=3cm]{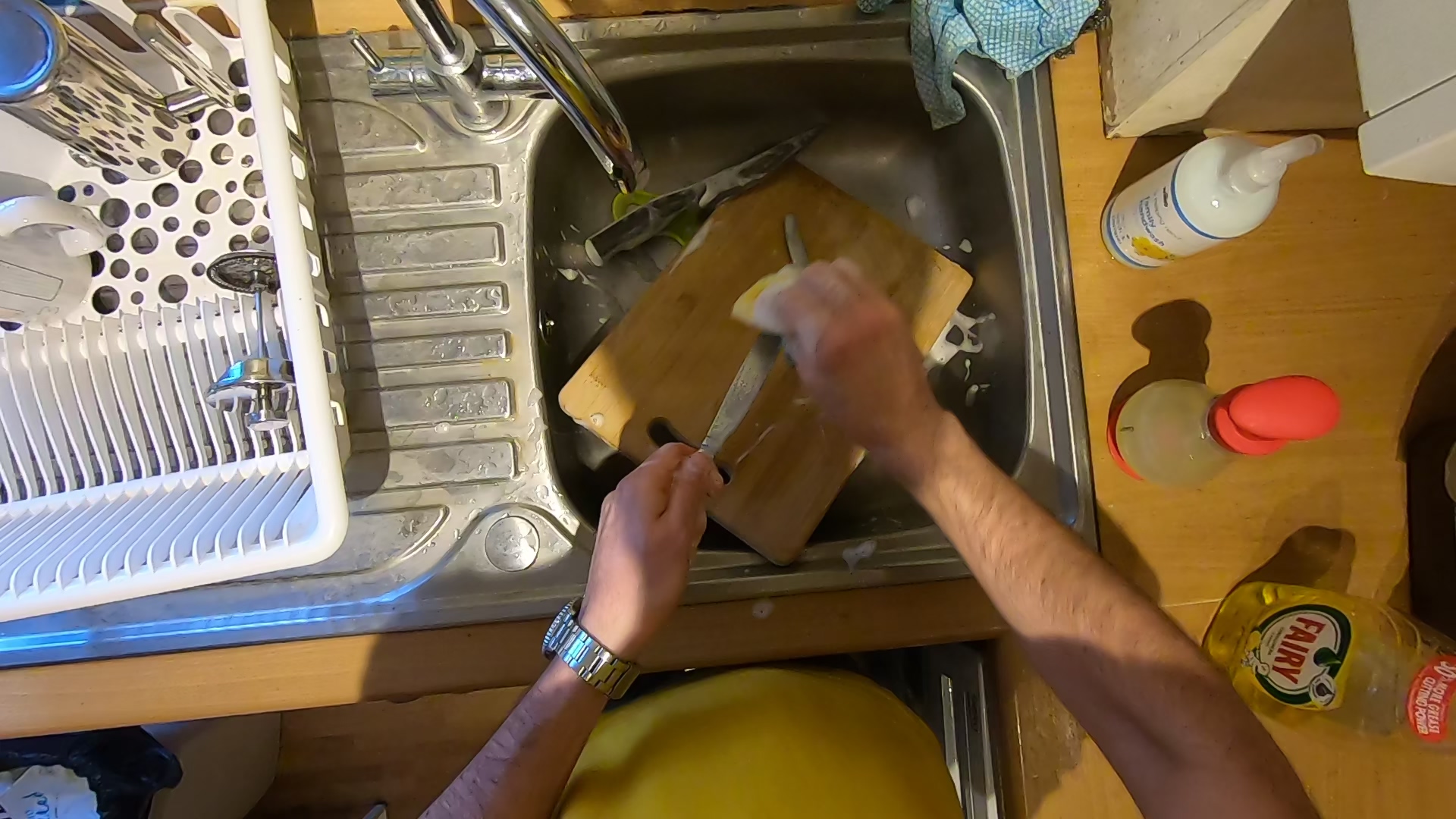}}  \\
\midrule
 
Kitchens \cite{epic_kitchen} & Instance & \raisebox{-\height}{\includegraphics[width=2cm]{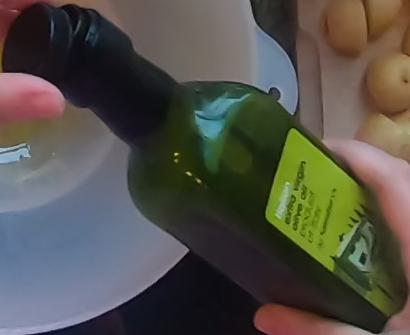}}  &  Given the bottle in the image, find an everyday image that contains the bottle.   & \raisebox{-\height}{\includegraphics[width=3cm]{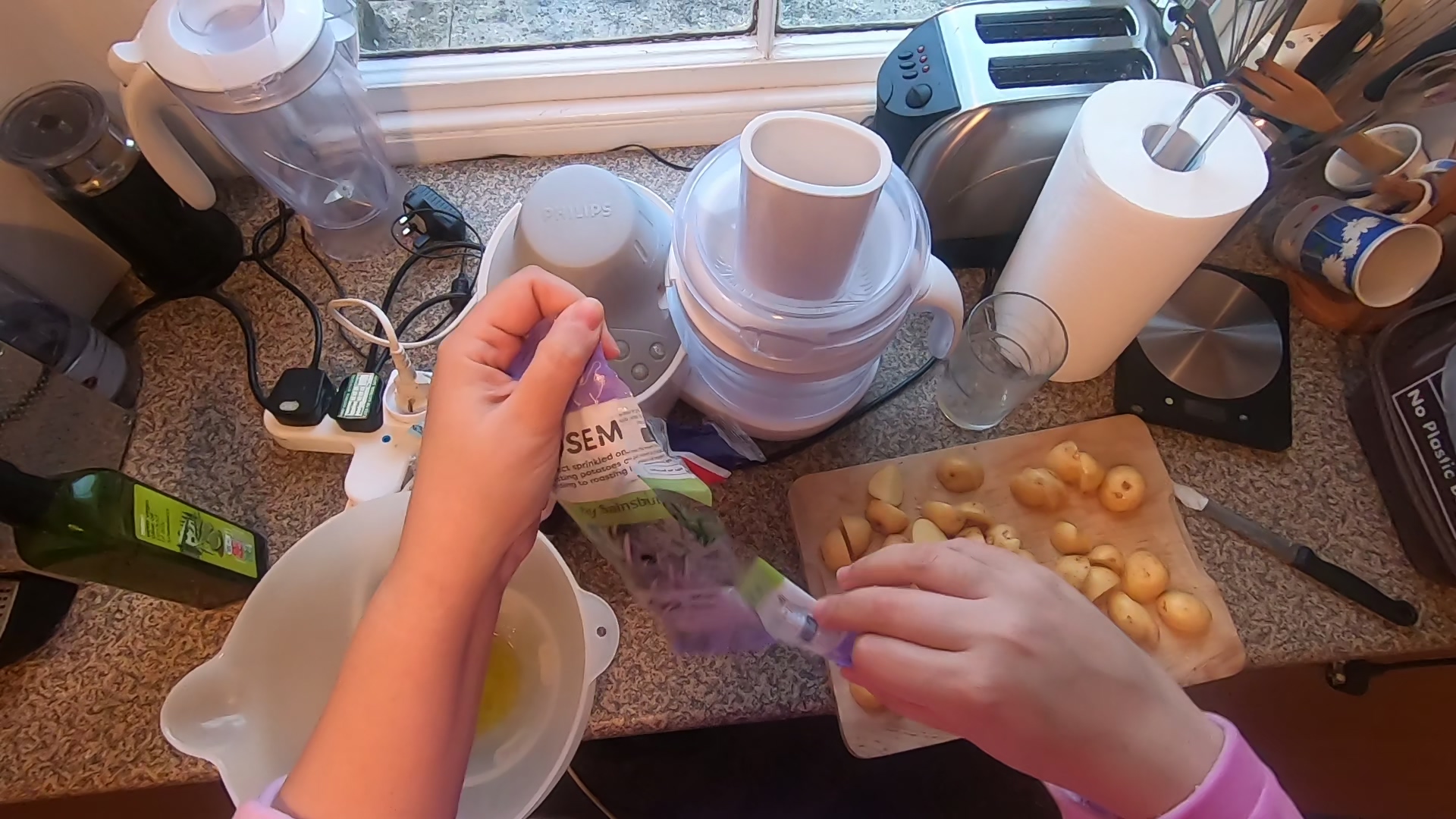}} \\
\midrule

Kitchens \cite{epic_kitchen}& Location & \raisebox{-\height}{\includegraphics[width=2cm]{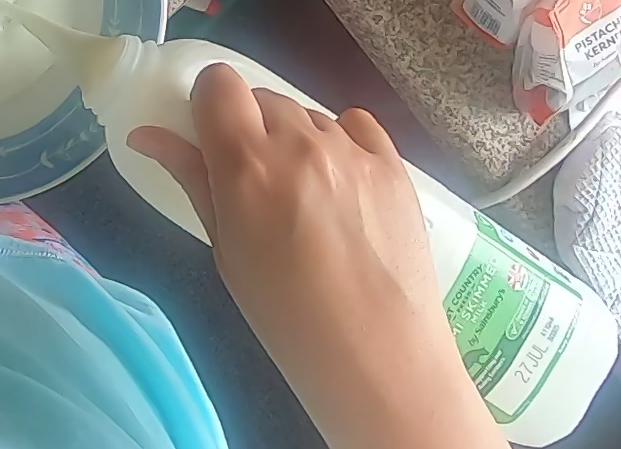}}  &   Find me an image containing the object in the given image with the following caption: The Bottle is located in the bottom right corner of the refrigerator door, next to a jar.    & \raisebox{-\height}{\includegraphics[width=3cm]{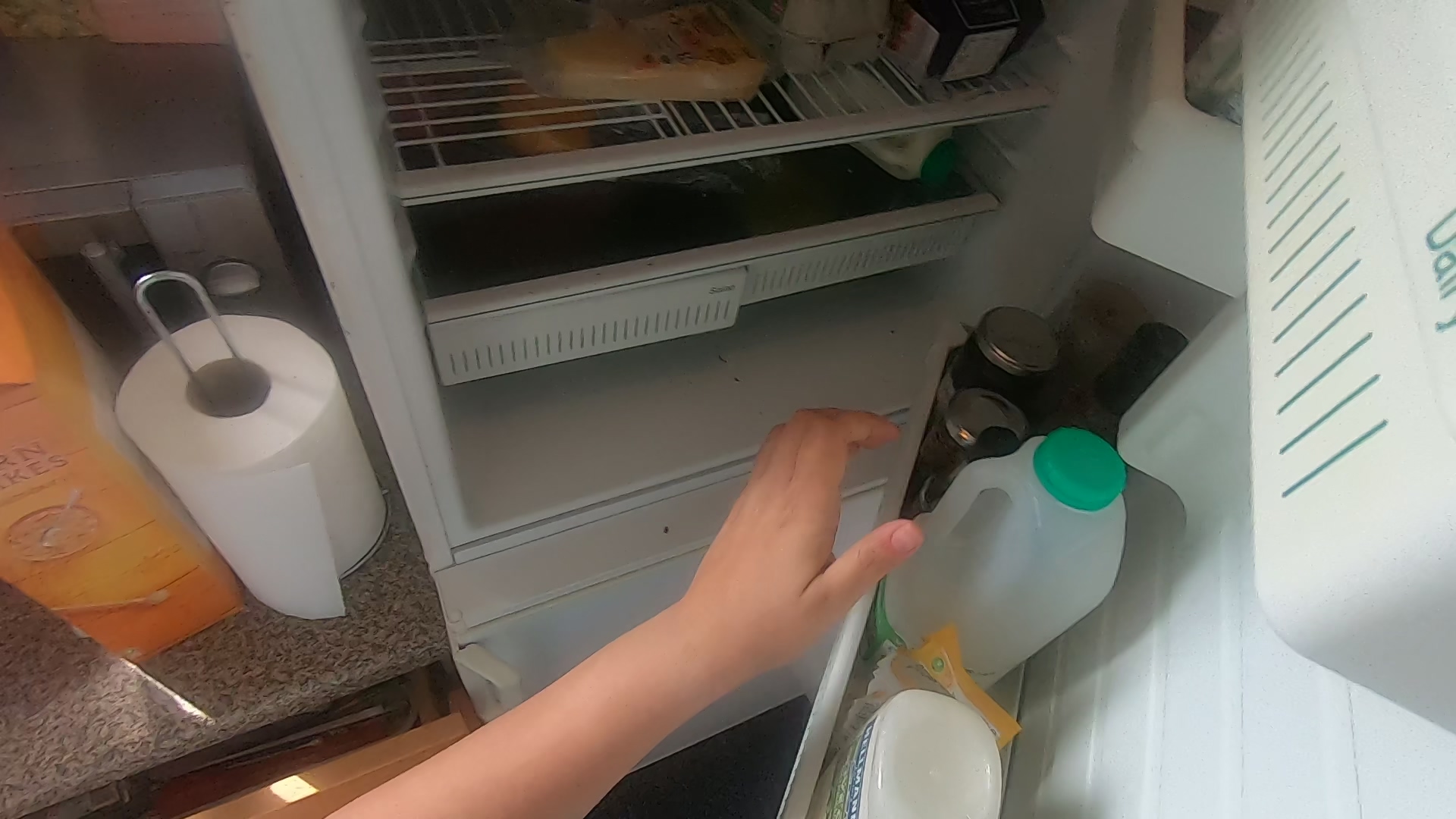}}  \\
 \midrule

Kitchens \cite{epic_kitchen} & Location & \raisebox{-\height}{\includegraphics[height=2cm]{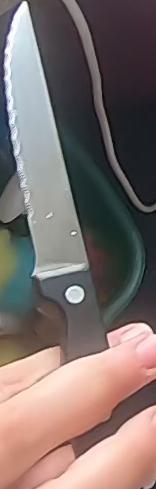}} &   Find me an image containing the object in the given image with the following caption: The Knife is located on a black cutting board, with a blue kitchen appliance to its right and a white juicer to its left.   & \raisebox{-\height}{\includegraphics[width=3cm]{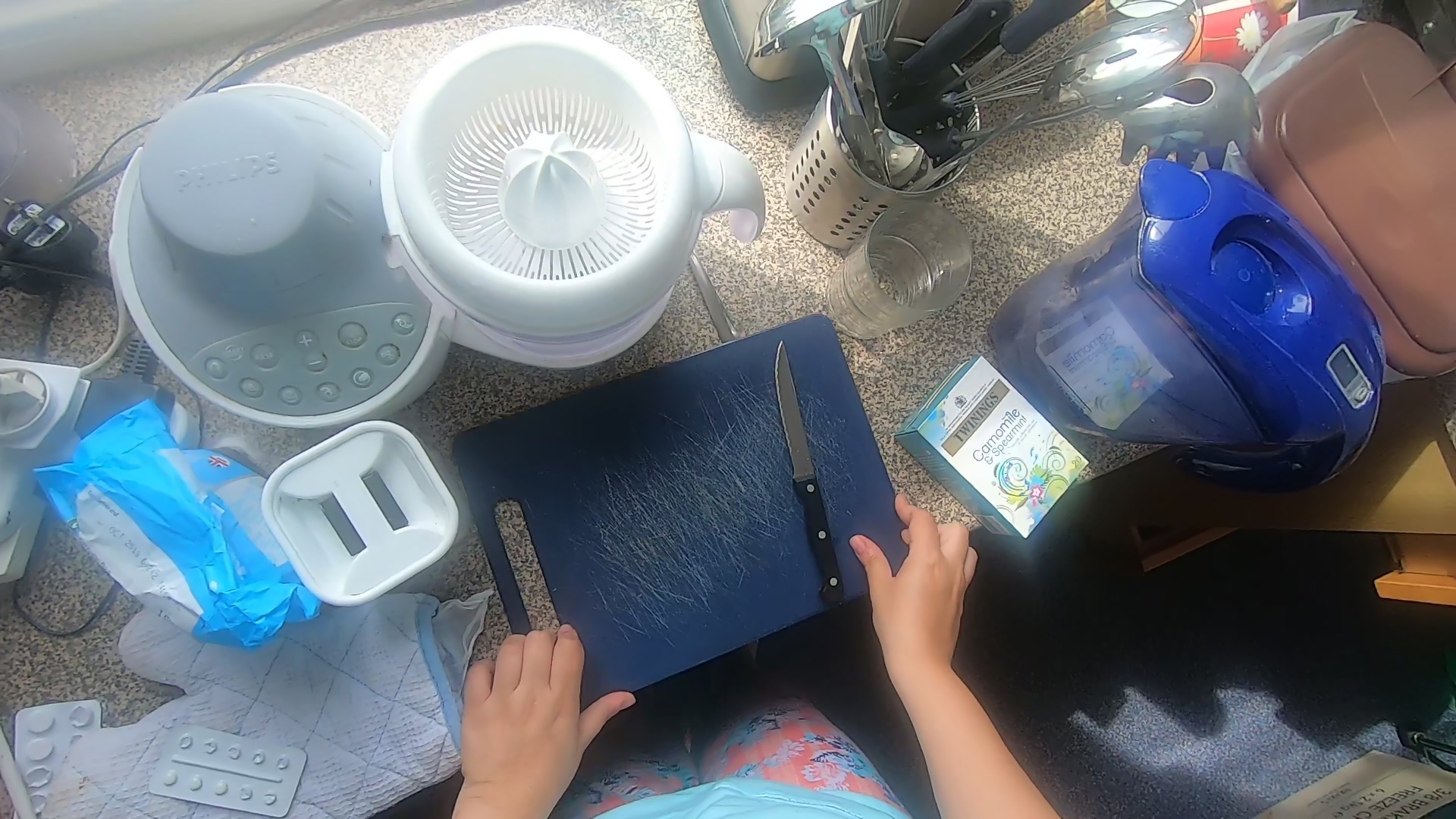}}  \\

\bottomrule
\end{tabular}
\caption{Examples of IDMR zero-shot testing dataset. }
\label{tab:visualize IDMR testing dataset}
\end{table*}

\begin{table*}[t!]
\centering
\begin{small}
\begin{tabular}{lccccc}
\hline
\multirow{2}{*}{Task} & \multicolumn{2}{c}{Fine-Tune} & \multicolumn{3}{c}{Train from MLLM} \\
\cline{2-3} \cline{4-6}
& VLM2Vec & MMRet & Ours(8B) & Ours(26B) & Ours(8B w/o. 557K synthetic data)\\
\hline
\multicolumn{6}{l}{\textbf{Classification (10 tasks)}} \\
\hline
ImageNet-1K & 65.6 & 58.8 & 70.5 & 80.6 & 70.7 \\
N24News & 79.5 & 71.3 & 80.2 & 81.6 & 79.6 \\
HatefulMemes & 67.1 & 53.7 & 72.9 & 72.3 & 70.5 \\
VOC2007 & 88.6 & 85.0 & 86.1 & 92.7 & 87.3 \\
SUN397 & 72.7 & 70.0 & 77.3 & 78.8 & 0.78 \\
\rowcolor{blue!10}
Place365 & 42.6 & 43.0 & 44.2 & 38.9 & 44.0 \\
\rowcolor{blue!10}
ImageNet-A & 19.3 & 36.1 & 39.3 & 63.6 & 36.3 \\
\rowcolor{blue!10}
ImageNet-R & 70.2 & 71.6 & 71.6 & 84.0 & 72.2 \\
\rowcolor{blue!10}
ObjectNet & 29.5 & 55.8 & 26.2 & 50.5 & 32.3 \\
\rowcolor{blue!10}
Country-211 & 13.0 & 14.7 & 14.7 & 20.3 & 14.7 \\
\rowcolor{gray!25}
All Classification &54.8 & 56.0 & 58.3 & 66.33 & 58.6 \\
\hline
\multicolumn{6}{l}{\textbf{VQA (10 tasks)}} \\
\hline
OK-VQA &  63.2 & 73.3 & 68.9 & 71.0 & 69.1 \\
A-OKVQA  & 50.2 & 56.7 & 56.6 & 59.2 & 56.8 \\
DocVQA & 78.4 & 78.5 & 73.0 & 75.1 & 72.1 \\
InfographicsVQA  & 40.8 & 39.3 & 40.9 & 44.6 & 43.3 \\
ChartQA & 59.0 & 41.7 & 62.9 & 64.6 & 61.0 \\
Visual7W & 47.7 & 49.5 & 52.1 & 54.9 & 51.6 \\
\rowcolor{blue!10}
ScienceQA & 43.4 & 45.2 & 52.1 & 54.7 & 53.8 \\
\rowcolor{blue!10}
VizWiz & 39.2 & 51.7 & 44.6 & 47.1 & 45.5 \\
\rowcolor{blue!10}
GQA & 60.7 & 59.0 & 61.2 & 71.0 & 58.8 \\
\rowcolor{blue!10}
TextVQA & 66.1 & 79.0 & 73.5 & 77.0 & 74.6 \\
\rowcolor{gray!25}
All VQA &54.9 & 57.4 & 58.6 & 61.9 & 58.7 \\
\hline
\multicolumn{6}{l}{\textbf{Retrieval (12 tasks)}} \\
\hline
VisDial & 73.3 & 83.0 & 80.7 & 81.5 & 80.5 \\
CIRR & 47.8 & 61.4 & 54.0 & 57.6 & 55.6 \\
VisualNews\textsubscript{t2i} & 67.2 & 74.2 & 73.3 & 78.5 & 74.0 \\
VisualNews\textsubscript{i2t} & 70.7 & 78.1 & 76.9 & 80.6 & 76.6 \\
MSCOCO\textsubscript{t2i}  & 70.6 & 78.6 & 76.9 & 79.1 & 75.8 \\
MSCOCO\textsubscript{i2t} & 66.5 & 72.4 & 73.7 & 75.4 & 73.5 \\
NIGHTS & 66.1 & 68.3 & 67.9 & 68.6 & 67.0 \\
WebQA  & 88.1 & 90.2 & 89.6 & 89.0 & 89.2 \\
\rowcolor{blue!10}
FashionIQ & 12.9 & 54.9 & 20.6 & 21.0 & 23.7 \\
\rowcolor{blue!10}
Wiki-SS-NQ & 56.6 & 24.9 & 64.0 & 66.9 & 65.2 \\
\rowcolor{blue!10}
OVEN & 47.3 & 87.5 & 58.2 & 67.4 & 58.3 \\
\rowcolor{blue!10}
EDIS & 79.9 & 65.6 & 88.7 & 87.6 & 88.6 \\
\rowcolor{gray!25}
All Retrieval &62.3 & 69.9 & 68.7 & 71.1 & 69.0 \\
\hline
\multicolumn{6}{l}{\textbf{Visual Grounding (4 tasks)}} \\
\hline
MSCOCO  & 67.3 & 76.8 & 75.2 & 81.5 & 71.6 \\
\rowcolor{blue!10}
RefCOCO  & 84.7 & 89.8 & 89.3 & 91.7 & 86.7 \\
\rowcolor{blue!10}
RefCOCO-matching & 79.2 & 90.6 & 88.1 & 88.1 & 86.9 \\
\rowcolor{blue!10}
Visual7W-pointing  & 86.8 & 77.0 & 89.9 & 93.1 & 84.1 \\
\rowcolor{gray!25}
All Visual Grounding &79.5 & 83.6 & 85.6 & 88.6 & 82.3 \\
\hline
\multicolumn{6}{l}{\textbf{Final Score (36 tasks)}} \\
All IND & 66.5& 59.1 & 70.5 & 73.4 & 70.2 \\
\rowcolor{blue!10}
All OOD &52.0 & 68.0 & 57.9 & 63.4 & 57.9 \\
\rowcolor{gray!25}
All &60.1 & 64.1 & 64.9 & 69.2 & 64.7 \\

\hline
\end{tabular}
\end{small}
\caption{Performance of each task on MMEB. }
\label{tab:MMEB full results}
\end{table*}


\end{document}